\documentclass{article}

\usepackage{PRIMEarxiv}

\usepackage[utf8]{inputenc} % allow utf-8 input
\usepackage[T1]{fontenc}    % use 8-bit T1 fonts
\usepackage{hyperref}       % hyperlinks
\usepackage{url}            % simple URL typesetting
\usepackage{booktabs}       % professional-quality tables
\usepackage{amsfonts}       % blackboard math symbols
\usepackage{nicefrac}       % compact symbols for 1/2, etc.
\usepackage{microtype}      % microtypography
\usepackage{lipsum}
\usepackage{fancyhdr}       % header
\usepackage{graphicx}       % graphics
\graphicspath{{media/}}     % organize your images and other figures under media/ folder

\usepackage{etoolbox}
\newbool{NumberedSections}
\booltrue{NumberedSections}
\newcommand{\sectionPlus}[1]{
    \ifbool{NumberedSections}{
        \section{#1}}{
        \section*{#1}}
}
\newcommand{\subsectionPlus}[1]{
    \ifbool{NumberedSections}{
        \subsection{#1}}{
        \subsection*{#1}}
}
\newcommand{\subsubsectionPlus}[1]{
    \ifbool{NumberedSections}{
        \subsubsection{#1}}{
        \subsubsection*{#1}}
}

\title{Composite Bayesian Optimization In Function Spaces Using NEON - Neural Epistemic Operator Networks
}

\author{
  Leonardo Ferreira Guilhoto \\
  Graduate Group in Applied Mathematics and Computational Science \\
  University of Pennsylvania \\
  Philadelphia, PA 19104 - US\\
  \texttt{guilhoto@sas.upenn.edu} \\
   \And
  Paris Perdikaris \\
  Department of Mechanical Engineering and Applied Mechanics \\
  University of Pennsylvania \\
  Philadelphia, PA 19104 - US\\
  \texttt{pgp@seas.upenn.edu} \\
}

\begin{document}

\maketitle

\begin{abstract}
    Operator learning is a rising field of scientific computing where inputs or outputs of a machine learning model are functions defined in infinite-dimensional spaces. In this paper, we introduce NEON (Neural Epistemic Operator Networks), an architecture for generating predictions with uncertainty using a single operator network backbone, which presents orders of magnitude less trainable parameters than deep ensembles of comparable performance. We showcase the utility of this method for sequential decision-making by examining the problem of composite Bayesian Optimization (BO), where we aim to optimize a function $f=g\circ h$, where $h:X\to C(\mathcal{Y},\R^{d_s})$ is an unknown map which outputs elements of a function space, and $g: C(\mathcal{Y},\R^{d_s})\to \R$ is a known and cheap-to-compute functional. By comparing our approach to other state-of-the-art methods on toy and real world scenarios, we demonstrate that NEON achieves state-of-the-art performance while requiring orders of magnitude less trainable parameters.
\end{abstract}

% keywords can be removed

\sectionPlus{Introduction}

 High-dimensional problems are prominent across all corners of science and industrial applications. Within this realm, optimizing black-box functions and operators can be computationally expensive and require large amounts of hard-to-obtain data for training surrogate models. Uncertainty quantification becomes a key element in this setting, as the ability to quantify what a surrogate model does not know offers a guiding principle for new data acquisition. However, existing methods for surrogate modeling with built-in uncertainty quantification, such as Gaussian Processes (GPs) \cite{RasmussenW06_gps4ml}, have demonstrated difficulty in modeling problems that exist in high dimensions. While other methods such as Bayesian neural networks \cite{bnn-survey-2020} (BNNs) and deep ensembles \cite{deep-ensambles-2016} are able to mitigate this issue, their computational cost can still be prohibitive for some applications. This problem becomes more prominent in Operator Learning, where either inputs or outputs of a model are functions residing in infinite-dimensional function spaces. The field of Operator Learning has had many advances in recent years\cite{Lu_2021_deeponet,li2021_FNO, kovachki2023neural_operator, kissas2022_LOCA, wang2021learning, wang2022improved}, with applications across many domains in the natural sciences and engineering, but so far its integration with uncertainty quantification is limited \cite{yang2022scalable,psaros2023uncertainty}.

In addition to safety-critical problems using deep learning such as ones in medicine \cite{filos2019systematic,Esteva2017dermatologist} and autonomous driving \cite{huang2020autonomous}, the generation of uncertainty measures can also be important for decision making when collecting new data in the physical sciences. Total uncertainty is often made up of two distinct parts: epistemic and aleatoric uncertainty. While epistemic uncertainty relates to ambiguity in a model due to lack of sufficient data or undersampling in certain regions of the domain, aleatoric uncertainty relates to ambiguity in the data itself, which can be noisy or come from processes that yield different outputs for the same input. In particular, when sequentially collecting new data, being able to quantify uncertainty and to decompose it into aleatoric and epistemic components can be of great importance\cite{BALD-2011, batchbald-2019,epistemicNNs}. Bayesian Optimization\cite{wang2022boreview, botorch} (BO), for example, iteratively collects new points of a hard-to-compute black-box function in order to maximize/minimize its value. The black-box nature of BO creates the need for accurate quantification of uncertainty in order to balance out exploration and exploitation of the design space. In order to determine the next best point to acquire, it is first needed to understand where the uncertainties of our system lie, and know what types of uncertainties drive optimal strategies.

A recent paradigm for studying uncertainty-producing models in deep learning is the framework of Epistemic Neural Networks (ENNs) \cite{epistemicNNs}. The ENN family  encompasses virtually any model which has built-in properties to estimate epistemic uncertainty, such BNNs \cite{bnn-survey-2020}, deep ensembles \cite{deep-ensambles-2016} and MCMC dropout \cite{gal2016dropout}. In order to model a function $f:X\to Y$, an ENN takes as input a feature vector $x\in X$ and an epistemic index $z\in \mathcal{Z}$, which is sampled at random from a distribution $P_Z$ independently of the input $x$. An ENN with parameters $\theta$ then outputs prediction $f_\theta(x,z)\in Y$. Since the variable $z$ has no connection to the data, it can be varied in order to produce an ensemble of predictions $f(x,z_1),\dots,f(x,z_n)$ for the same input $x$, where $z_1,\dots,z_n\overset{\text{\textit{iid}}}{\sim}P_Z$. This allows us to analyze statistics of these predictions in order to quantify epistemic uncertainty.

In the light of these advances, we propose integrating the ENN framework to operator learning problems in order to understand epistemic uncertainty in this setting. In addition, we also combine the EpiNet architecture\cite{epistemicNNs} with deep operator network models\cite{seidman2022nomad,Lu_2021_deeponet} to implement NEON, a computationally cheap alternative to deep ensembles of operator networks, which is the state-of-the-art approach to uncertainty quantification and operator composite Bayesian Optimization \cite{BHOURI2023116428}. Not only does this approach generally require significantly less trainable parameters than deep ensembles, it can be implemented as an augmentation to any previously trained model at a very cheap computational cost, leading to flexibility for many problems that require large neural operator models to get accurate results. Finally, EpiNets have been shown to excel at producing well-calibrated joint predictions\cite{epistemicNNs}, which is crucial for decision making processes.

In the light of these challenges, the contributions of this paper are as follows:
\begin{itemize}
    \item We propose \textbf{NEON (Neural Epistemic Operator Networks)}, a method for quantifying epistemic uncertainty in operator learning models. This architecture features flexibility for implementation on previously trained models and enables the quantification of epistemic uncertainty using a single model, removing the necessity to train large ensembles.
    \item We study different choices of acquisition functions for composite BO using the Epistemic Neural Network (ENN) framework. We additionally propose the \textbf{Leaky Expected Improvement} (L-EI) acquisition function, a variant of Expected Improvement (EI) that allows for easier optimization while retaining similar local extrema.
    \item By examining different benchmarks, we demonstrate NEON achieves \textbf{state-of-the-art performance} high-dimensional composite BO while often consisting of \textbf{1-2 orders of magnitude less training parameters}. This is consistent across different benchmarks when compared with the available literature\cite{BHOURI2023116428, maddox2021bayesian}.
\end{itemize}

\sectionPlus{Methods}

\subsectionPlus{Operator Learning for Composite Bayesian Optimization}

\textbf{Operator learning}\cite{Lu_2021_deeponet,chenchen1995,li2021_FNO,kovachki2023neural_operator} is a field of scientific machine learning that aims to learn maps between functional spaces using labeled data. Formally, we let $C(\mathcal{X}, \R^{d})$ denote the set of continuous functions from $\mathcal{X}$ to $\R^d$ and, following the notation of \cite{seidman2022nomad}, if $\mathcal{X}$ and $\mathcal{Y}$ are compact spaces, $d_u,d_s\in\N$ and $\mathcal{G}:C(\mathcal{X}, \R^{d_u})\to C(\mathcal{Y}, \R^{d_s})$ is an operator between function spaces, the goal of operator learning is to learn the behaviour of $\mathcal{G}$ from observations $\{(u_i,s_i)\}_{i=1}^N$ where $u_i \in C(\mathcal{X}, \R^{d_u})$ and $s_i = \mathcal{G}(u_i)\in C(\mathcal{Y}, \R^{d_s})$. This is done by fitting a model $\mathcal{F}_\theta$ with trainable parameters $\theta\in\Theta\subseteq \R^{d_{\theta}}$ minimizing the empirical risk
\begin{equation}
    \mathcal{L}(\theta) = \frac{1}{N}\sum_{i=1}^N || \mathcal{F}_\theta(u_i) - s_i ||_{L^2} = \frac{1}{N}\sum_{i=1}^N || \mathcal{F}_\theta(u_i) - \mathcal{G}(u_i) ||_{L^2}^2.
\end{equation}
It is also common to fit the model minimizing the \textit{relative} empirical risk\cite{di2023neural,wang2022improved},
\begin{equation}
    \mathcal{L}_{\text{relative}}(\theta) = \frac{1}{N}\sum_{i=1}^N \frac{|| \mathcal{F}_\theta(u_i) - s_i ||_{L^2}}{|| s_i ||_{L^2}} = \frac{1}{N}\sum_{i=1}^N \frac{|| \mathcal{F}_\theta(u_i) - \mathcal{G}(u_i) ||_{L^2}^2}{||\mathcal{G}(u_i) ||_{L^2}^2},
\end{equation}
which is the approach we take in the experiments carried out in this paper.

In many cases (including the ones considered in our experiments), the input to the operator $\mathcal{G}$ can also be a finite-dimensional vector $u\in X\subseteq \R^{d_u}$ instead of a proper function. Using the operator learning framework in this scenario, however, can still be advantageous, as existing architectures such as the DeepONet\cite{Lu_2021_deeponet, chenchen1995}, LOCA\cite{kissas2022_LOCA}, MIONet\cite{jin2022mionet} and more, allow for efficient evaluation of the output function $s=\mathcal{G}(u)\in C(\mathcal{Y}, \R^{d_s})$ at any desired query point $y\in \mathcal{Y}$ instead of being restricted to a fixed grid.

\begin{wrapfigure}{R}{0.5\textwidth}
\centering
\includegraphics[width=0.5\textwidth]{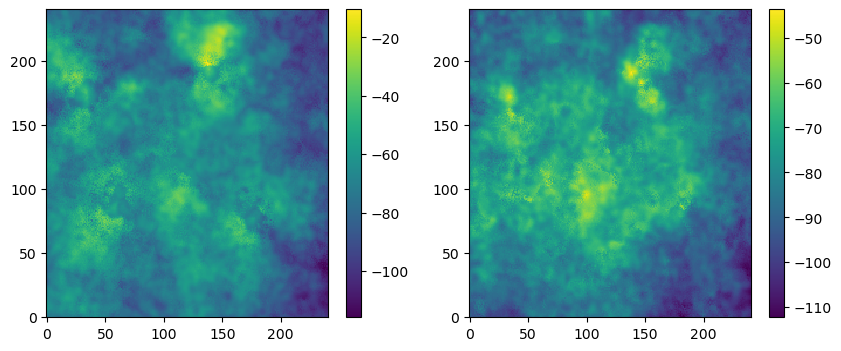}
\caption{\label{fig:cell_plots} Example of $h(u)\in C([0,221]^2,\R^2)$ for the the Cell Towers problem. The input $u\in\R^{30}$ encodes transmission parameters of 15 cell towers, which are used to produce the function seen above, where signal intensity and interference are plotted, respectively. This information is the used to compute a score $f(u)=g(h(u))\in\R$ which evaluates the quality of cellular service in the region. By using operator composite BO, we take advantage of the known compositional structure of $f=g\circ h$, and only need to model the behaviour of $h$.} \end{wrapfigure}

\textbf{Bayesian Optimization (BO)} consists of maximizing a black-box function $f:X\to\R$ where we do not have access to the gradients of $f$, but only evaluations $f(x_1),\dots f(x_n)\in \R$ at a given set of points $x_1,\dots,x_n\in X$. The goal of BO is then to determine $\argmax_{x\in X}f(x)$ by iteratively evaluating $f$ at new points with the fewest possible calls to the function $f$. In order to determine what points to evaluate $f$ at next, we first fit a \textit{surrogate model} to the data $\{\left(x_i,f(x_i)\right)\}_{i=1}^N$ which gives us a probability measure $P$ over the possible functions $\hat{f} : X\to\R$ that agree with the observations $\{\left(x_i,f(x_i)\right)\}_{i=1}^N$. Using this probability measure, we then define an \textit{acquisition function} $\alpha:X\to \R$ and set $x_{N+1}=\argmax_{x\in X}\alpha(x)$. Intuitively, one can think of $\alpha$ as a function that quantifies the utility of evaluating $f$ at a given point. There are many possible choices for acquisition functions in BO, which have different approaches to balancing out exploitation (evaluating $f$ at points we believe $f$ is high) and exploration (evaluating $f$ at points where there is little information available). Some of these acquisition functions are discussed later in this manuscript.

Although traditional BO already presents its fair share of challenges and open problems, in this paper we consider \textbf{Composite Bayesian Optimization} \cite{astudillo2019bayesian, BHOURI2023116428, maddox2021bayesian, maus2023joint}, where the function $f:X\to\R$ we wish to optimize is the composition $f=g\circ h$ of two functions $h:X\to Y$ and $g:Y\to \R$. In this setting, we also assume that $h$ is a hard to compute black-box function, while $g$ is a known map which can be computed exactly and cheaply. This setting is common in engineering applications, where may $h$ represent a complex physical process which is hard to evaluate (such as the solution of a PDE, or the evolution of a complicated dynamical system), and $g$ is a simple function (such as a directly computable property of a dynamical system like temperature). Composite BO is of particular importance to problems in the field of optimal control, and recent research has shown that this approach outperforms vanilla BO \cite{maddox2021bayesian, BHOURI2023116428, astudillo2019bayesian}, as in the latter approach only the direct function $f$ is taken into account (ignoring information about composite structure of $f$ and the known map $g$).

In this paper, we consider composite BO cases where $X\subseteq \R^{d_u}$ is a finite-dimensional space and $Y=C(\mathcal{Y},\R^{d_s})$ is a space of continuous functions.  That is, $h:X\to C(\mathcal{Y},\R^{d_s})$ is a complex function for which obtaining data is expensive and $g:C(\mathcal{Y},\R^{d_s})\to\R$ is a known and cheap-to-compute functional. As an example, consider the problem of optimizing the coverage area and interference of the service provided by a set of cell towers\cite{dreifuerst2021optimizing}. In this case, $u\in\R^{30}$ encodes possible configurations of antenna transmission powers and down-tilt angles of 15 different cell towers, while $h(u)\in C(\R^2, \R^2)$ is the function that outputs the signal power and interference strength, respectively, at any desired location in $R^2$. An example plot for signal strength and interference intensity is shown in Figure \ref{fig:cell_plots}. Finally, $f(u)=g(h(u))\in\R$ uses this map to compute a score that evaluates the quality of coverage and extent of interference between towers. In this scenario, predicting the map $h(u)$ of signal strength/interference is a challenging problem involving complex physics, while computing $g$ given this map is a simple and computationally cheap task. For problems like this, it is possible to use operator learning to infer the map $h$ from data and use composite BO to determine the best configuration $u$ for antenna powers and down-tilt angles. We explore this particular problem, among others, in our experiments section.

The choice of using traditional or composite BO is problem dependent and entails different trade-offs. If no compositional structure is known, and $f$ is completely black-box, traditional BO is the only possible approach. However, when information about the structure of $f$ allows for the formulation $f=g\circ h$ in the manner prescribed by the composite BO framework, studies show that the composite BO formulation achieves better results and faster convergence\cite{astudillo2019bayesian, kim2022deep}.

\begin{table}[ht]
\centering
\begin{tabular}{|c|c|c|c|}
\hline
Paradigm & Maps & Known & Unknown - Must be Modeled\\
\hline
Traditional BO & $X \xrightarrow{\ \ f\ \ } \R$ & N/A & $f$\\
\hline
Composite BO & $X \xrightarrow{\ \ h\ \ } Y \xrightarrow{\ \ g\ \ } \R $ & $g$ & $h$\\
\hline
Operator Composite BO & $X \xrightarrow{\ \ h\ \ } C(\mathcal{Y}, \R^{d_s}) \xrightarrow{\ \ g\ \ } \R $ & $g$ & $h$\\
\hline
\end{tabular}
\caption{\label{tab:bo-paradigms} Representation of the different BO paradigms, showing the relationship between the input space $X$ of $f:X\to\R$ and potentially known structures about $f$. For traditional (non-operator) composite BO, the intermediate space $Y\subseteq \R^{d_Y}$ is taken to be a subset of $\R^{d_Y}$ for some $d_Y \in \N$, while in operator composite BO the intermediate space $Y\subseteq C(\mathcal{Y}, \R^{d_s})$ is taken to be a space of continuous functions from $\mathcal{Y}$ to $\R^{d_s}$.}
\end{table}

\subsectionPlus{Epistemic Neural Networks as a Flexible Framework for Uncertainty-Aware Models}

The Epistemic Neural Networks (ENNs) \cite{epistemicNNs} framework is a flexible formal setup that describes many existing methods of uncertainty estimation such as Bayesian Neural Networks (BNNs) \cite{bnn-survey-2020}, dropout \cite{gal2016dropout} and Deep Ensembles \cite{deep-ensambles-2016}. An ENN with parameters $\theta\in\Theta\subseteq \R^{d_\theta}$ takes as input a feature vector $x\in\R^{d_x}$ along with a random index $z\in\R^{d_z}$ and outputs a prediction $f_\theta(x,z)$. Under this framework, the random index $z$ is sampled from probability distribution $P_Z$ independently from $x$ and therefore has no connection to the data and can be varied in order to obtain different predictions.

For example, a deep ensemble with $n$ networks falls into the ENN umbrella in the following manner. The random index $z$ can take values in the discrete set $\{1,2,\dots,n\}$, indicating which network we use to carry out the prediction. In this way, varying $z$ gives us variety of predictions which allow us to estimate uncertainty by checking how much they agree or disagree with each other. If enough data is available and the model is well trained, we expect the predictions to not depend strongly on $z$, meaning that there is little disagreement between different networks. On the other hand, if our predictions vary a lot depending on $z$, this indicates that there is large model (also called \textit{epistemic}) uncertainty. Another example of this are Bayesian neural networks (BNNs), where the weights of the network are sampled according to a re-parametrization trick using $z$, which in this case takes on continuous values. In \cite{epistemicNNs}, the authors prove that BNNs are a special case of the ENN framework.

Under this framework, we can use the same feature $x$ and vary $z$ in order to estimate the \textit{epistemic} uncertainty of our model, which is crucial in the context of Bayesian Optimization.

\subsubsectionPlus{EpiNets}

The paper \cite{epistemicNNs} also introduces \textit{EpiNets}, which implements an ENN in a simple and computationally cheap manner as
\begin{equation}
    \underbrace{f_\theta(x, z)}_{ENN} = \underbrace{\mu_\zeta(x)}_{\text{base net}} + \underbrace{\sigma_\eta\left(\texttt{sg}[\phi_\zeta(x)], z\right)}_{\text{EpiNet}}.
\end{equation}
Here, $\zeta$ is the parameters of the base network (which does not take $z$ into account), $\eta$ denotes the parameters of the EpiNet, $\theta=(\zeta, \eta)$ is the overall set of trainable parameters, $\texttt{sg}$ is a \textit{stop gradient} operation that blocks gradient back-propagation from the EpiNet to the base network, and $\phi_\zeta$ is a function that extracts useful features from the base network, such as the last layer of activations concatenated with the original input.

The EpiNet itself is composed of two parts, a learnable component $\sigma^L_\eta$, and a prior component $\sigma^P$ which is initialized randomly and is not optimized, 
\begin{equation}\label{eq:EpiNet}
    \underbrace{\sigma_\eta\left(\Tilde{x}, z\right)}_{\text{EpiNet}} = \underbrace{\sigma^L_\eta\left(\Tilde{x}, z\right)}_{\text{learnable}} + \underbrace{\alpha\sigma^P\left(\Tilde{x}, z\right)}_{\text{prior}},
\end{equation}
where $\alpha$ is a hyperparameter called the \textit{scale} of the prior. Feeding the random index $z$ along with the features ensures that $z$ does in fact play a role in predictions, as the learnable component of the EpiNet has to learn to counteract the effect of the untrained prior component as opposed to making only trivial zero predictions during training.

EpiNets readily enable the quantification of epistemic uncertainty for any given neural network architecture, including any pre-trained models that can be treated as a black-box. Even for large base models with tens of millions of parameters, augmentation by a small EpiNet has been demonstrated to yield good uncertainty estimates at little extra computational cost \cite{epistemicNNs}. This poses a significant advantage over deep ensembles, for example, where several models need to be trained. In addition, the EpiNet can be trained in conjunction to the base network, or at a later stage, after the base network has been trained and its parameters fixed. This yields further flexibility, as an EpiNet can be used to augment any pre-trained base network, even if it does not originally fit into the ENN umbrella.

\subsectionPlus{Epistemic Operator Networks as Surrogate Models}

Given an unknown function $h:X\to C(\mathcal{Y}, \R^{d_s})$, we assume an epistemic model, which, given inputs $u,y,z$ and parameters $\theta$, computes predictions $\hat{h}_\theta(u,y,z)\approx h(u)(y)\in \R^{d_s}$, where $u\in X\subseteq \R^{d_u}$ is a feature vector, $y\in\mathcal{Y}$ is a query point and $z\in\mathcal{Z}$ is a random epistemic index with probability distribution $P_Z$ over the space $\mathcal{Z}$. Given $u$ and $y$, the predictive distribution for the unknown quantity $h(u)(y)$ is equal to that of the transformed random variable $\hat{h}_\theta(u,y,Z)$ where $Z\sim P_Z$. That is, draws from the predictive distribution of $h(u)(y)$ can be obtained by first sampling $Z\sim P_Z$ and then computing the forward pass $\hat{h}_\theta(u,y,Z)$.

Alternatively, if the output of $\hat{h}_\theta(u,y,z)$ is a probability distribution itself (in cases such as ones using Negative Log-Likelihood (NLL) loss), instead of a single point in $\R^{d_s}$ we model our \textit{predictive distribution} for $h(u)(y)\in\R^{d_s}$ as a mixture of the output of the model over different values of $z$
\begin{equation}\label{eq:predictive-distribution}
    p_\theta(h(u)(y)|u,y) = \int p(h(u)(y)|u,y,z) P_Z(z)dz = \int \hat{h}_\theta(u,y,z) P_Z(z)dz = \mathbb{E}_z\left[\hat{h}_\theta(u, y ,z)\right].
\end{equation}
This implies our \textit{predictive distribution} over the objective function $f=g\circ h$ is
\begin{equation}\label{eq:predictive-distribution-objective}
    p_\theta(f(u)|u) = \int p(f(u)|u,z) P_Z(z)dz = \int g(\hat{h}_\theta(u,\cdot ,z)) P_Z(z)dz = \mathbb{E}_z\left[g\left(\hat{h}_\theta(u, \cdot ,z)\right)\right].
\end{equation}
Although we do not take this approach in the experiments considered in this paper, training a model in this manner further allows for disentangling aleatoric and epistemic uncertainty and is the potential subject of future research.

\subsectionPlus{NEON - An Architecture for Epistemic Operator Learning}

NEON (Neural Epistemic Operator Network) is a combination of the operator learning framework\cite{seidman2022nomad, Lu_2021_deeponet, li2021_FNO} with the EpiNet\cite{epistemicNNs} architecture. More specifically, it is a version of the EpiNet architecture where the base network is an operator learning model.

 \begin{figure}[ht!]
    \centering\includegraphics[width=.6\textwidth]{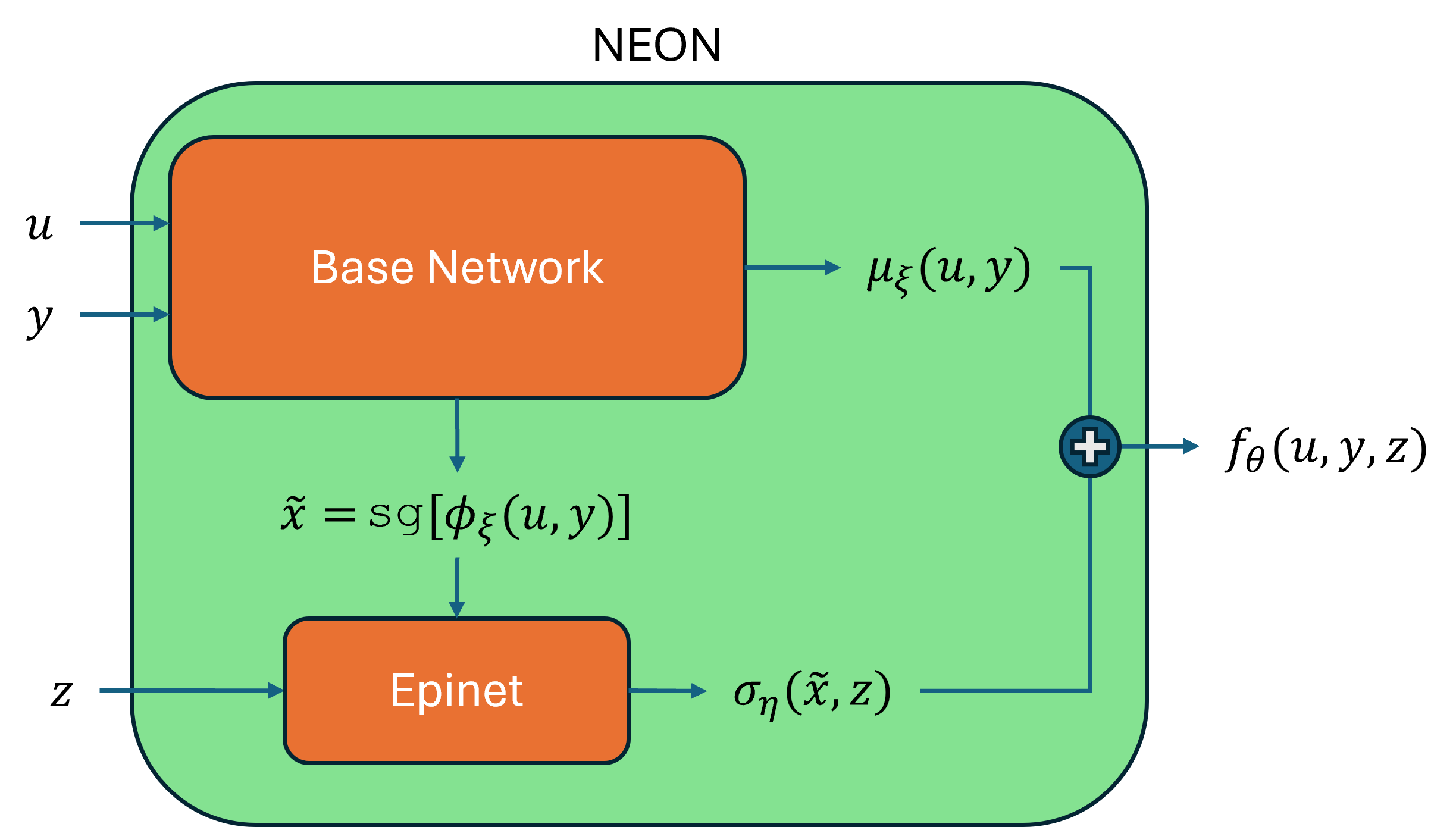}\hfill
    \\[\bigskipamount]
    \centering\includegraphics[width=.4\textwidth]{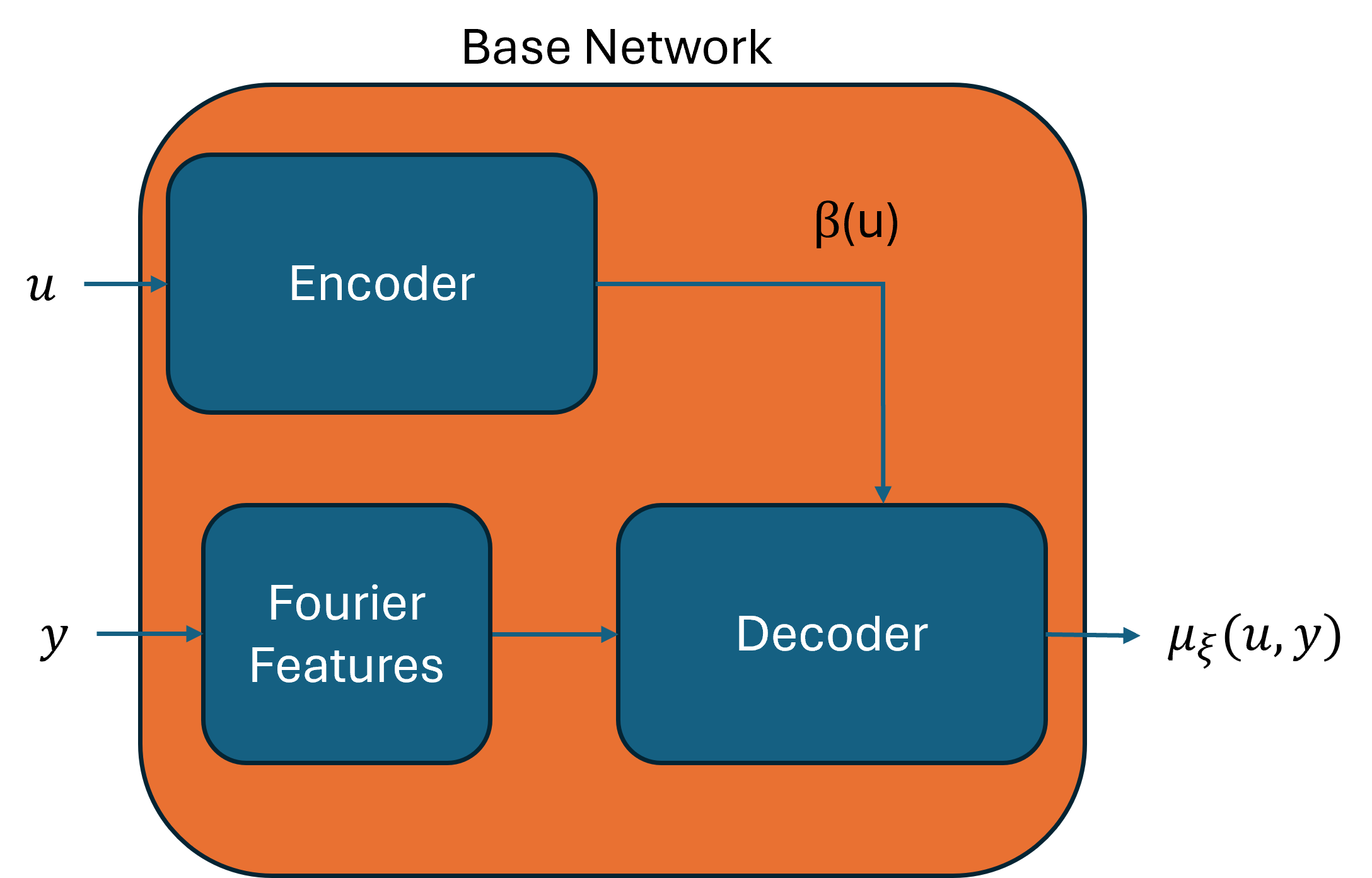}\hspace{0.05\textwidth}
    \includegraphics[width=.5\textwidth]{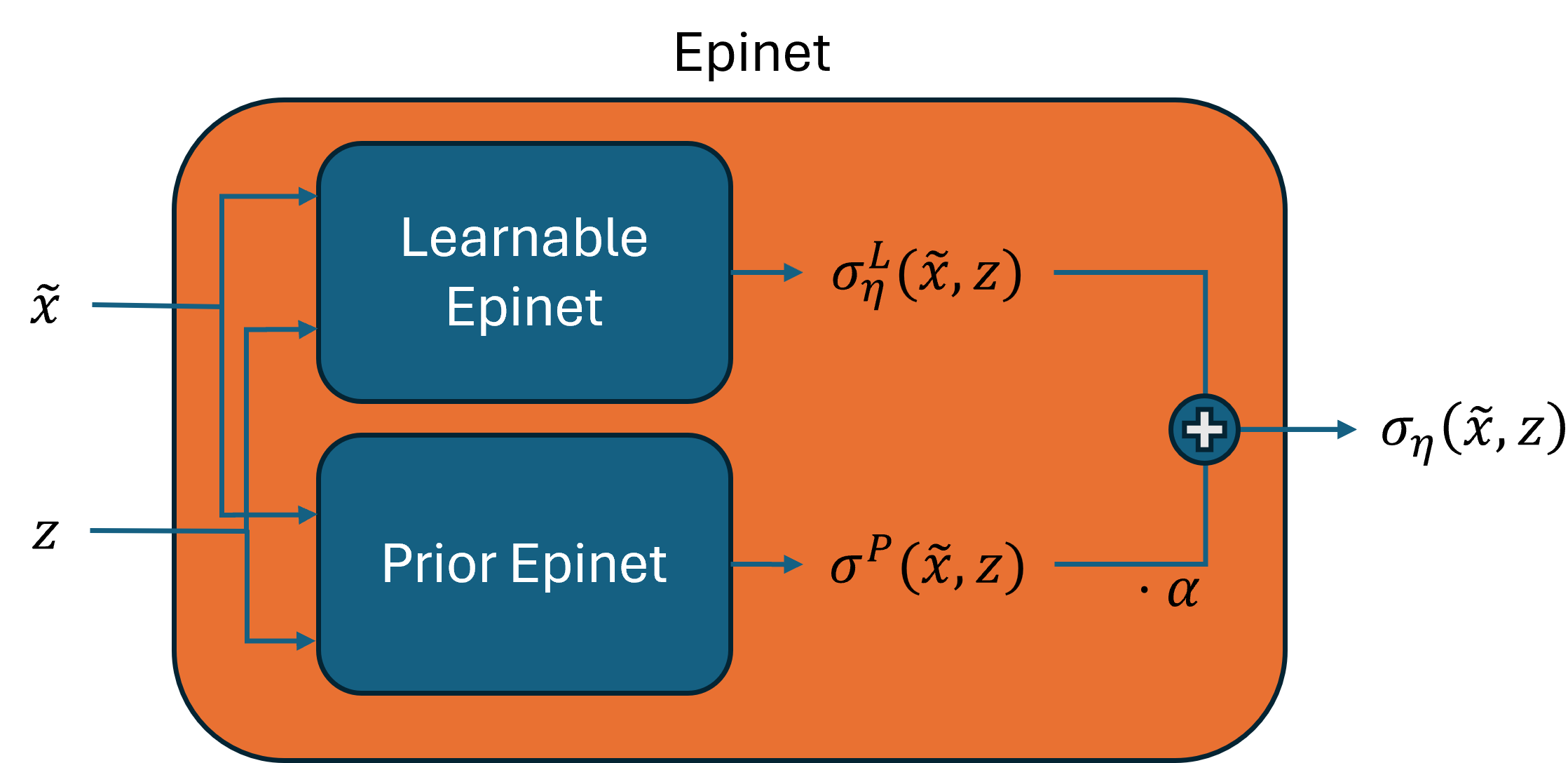}
    \caption{Diagrams for the architectures used in this paper. The NEON architecture (top) combines the deterministic output of the base network with the stochastic output of the small EpiNet in order to produce predictions $f_\theta(u,y,z)$. The base network (bottom left) uses an encoder/decoder structure, with the encoder being dependent only on $u$, while the decoder receives as input the latent representation $\beta(u)$ and a Fourier feature encoding of the query point $y$. Finally, the EpiNet (bottom right) receives as input features $\Tilde{x}=\texttt{sg}[\phi_\xi(u,y)]$ from the base network along with a random epistemic index $z\sim P_Z$. Here, \texttt{sg} denotes the ``stop gradient'' operation. The EpiNet is composed of a learnable component $\sigma^L_\eta$ that changes through training, and a prior network $\sigma^P$ that is not affected by the data.}\label{fig:network-diagrams}
\end{figure}

The operator learning architectures used as base networks in this paper follow a encoder/decoder structure, where the encoder $e_{\xi_e}:\R^{d_u}\to\R^{d_\beta}$ with parameters $\xi_e$ takes in as input $u\in \R^{d_u}$ and outputs a latent variable $\beta=e_{\xi_e}(u)\in\R^{d_\beta}$, and the decoder $d_{\xi_d}: \R^{d_\beta}\times \mathcal{Y}\to\R^{d_s}$ with parameters $\xi_d$ takes as input $\beta$ and a query point $y\in \mathcal{Y}$ and outputs the prediction $d_{\xi_d}(\beta, y)\in \R^{d_s}$. Thus, the base network $\mu_\xi$ makes predictions of the form $\mu_\xi(u, y) = d_{\xi_d}(\beta, y) = d_{\xi_d}(e_{\xi_e}(u), y)$, where $\xi = (\xi_e, \xi_d)$ are the trainable parameters. In the experiments considered in this paper, the encoder is a Multi-Layer Perceptron (MLP), while the decoder is either another MLP taking as input a concatenation of $\beta$ and a Fourier-Feature\cite{tancik2020fourfeat} encoding of $y$, or a Split Decoder, which is an architecture inspired by \cite{rebain2022attention} and further described below. Despite these specific choices, NEON can use any other choice of architecture for the backbone of the encoder or decoder.

As for the architectures of the EpiNets, we use MLPs, which take as input $\Tilde{x}=\texttt{sg}[\phi_\xi(u, y)]$ and $z$, where \texttt{sg} is the \textit{stop gradient} operation, which stops backpropagation from the EpiNet to the base network during training. In the architectures implemented in this paper, we take $\phi_\xi(u, y)$ to be the concatenation $\phi_\xi(u, y)=(\beta, \mu^{\text{last}}, y)$, where $\beta=e_{\xi_e}(u)$ is the output of the encoder, $\mu^{\text{last}}=\mu_\xi^{\text{last}}(u,y)$ is the last layer activations of the decoder, and $y\in \mathcal{Y}$ is the query point. Both the trainable and prior EpiNet networks then take as input the pair $(\Tilde{x}, z)$. This choice is similar to the original formulation of EpiNets in \cite{epistemicNNs}.

A diagram illustrating this choice of architecture can be seen in Figure \ref{fig:network-diagrams}, and details about hyperparameters used for each experiment can be seen in the appendix.

\subsubsectionPlus{Decoder Architectures}

\begin{figure}[ht!]
    \centering
    \includegraphics[width=0.41\textwidth]{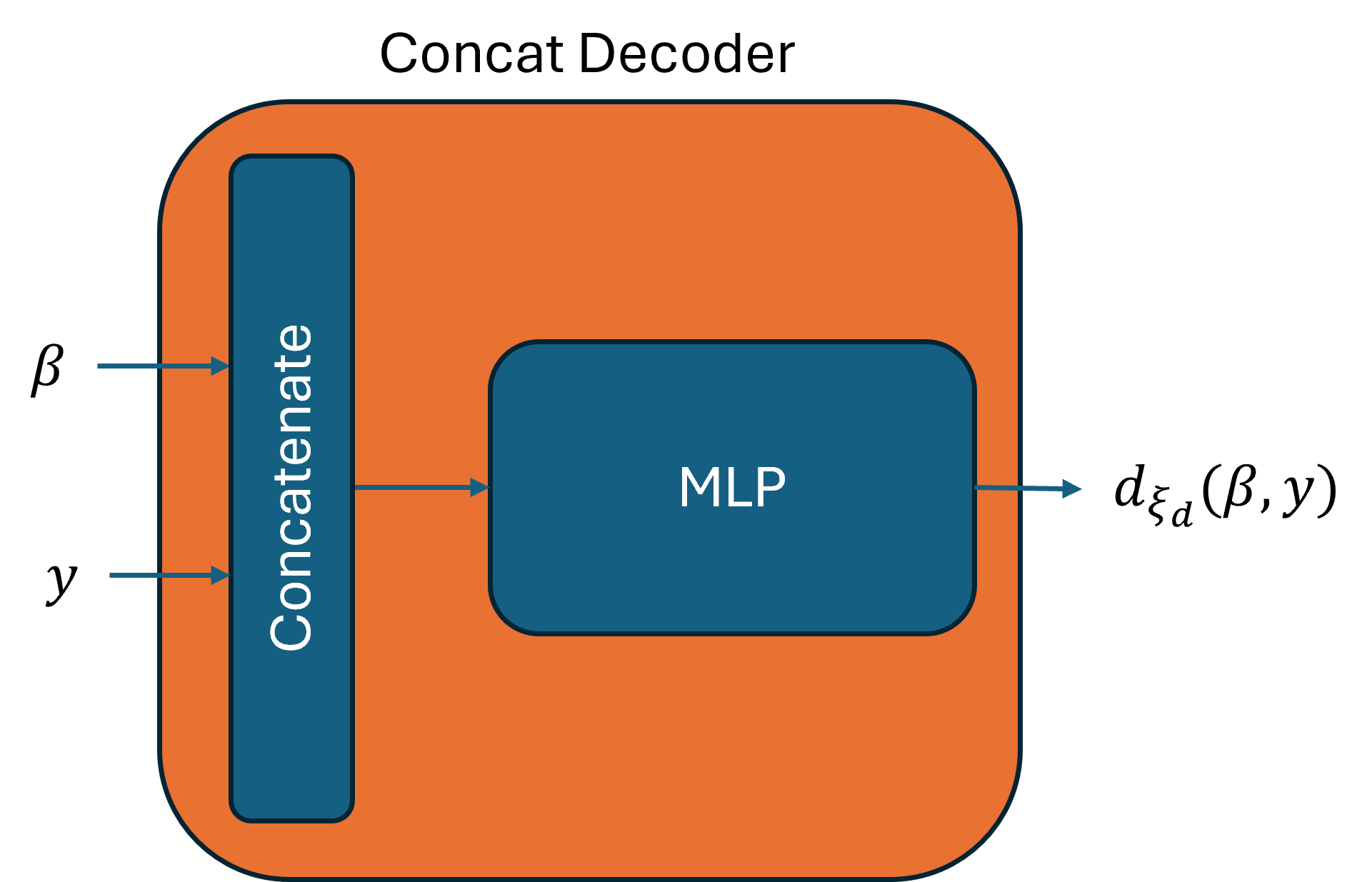}
    \hspace{0.01\textwidth}
    \includegraphics[width=0.56\textwidth]{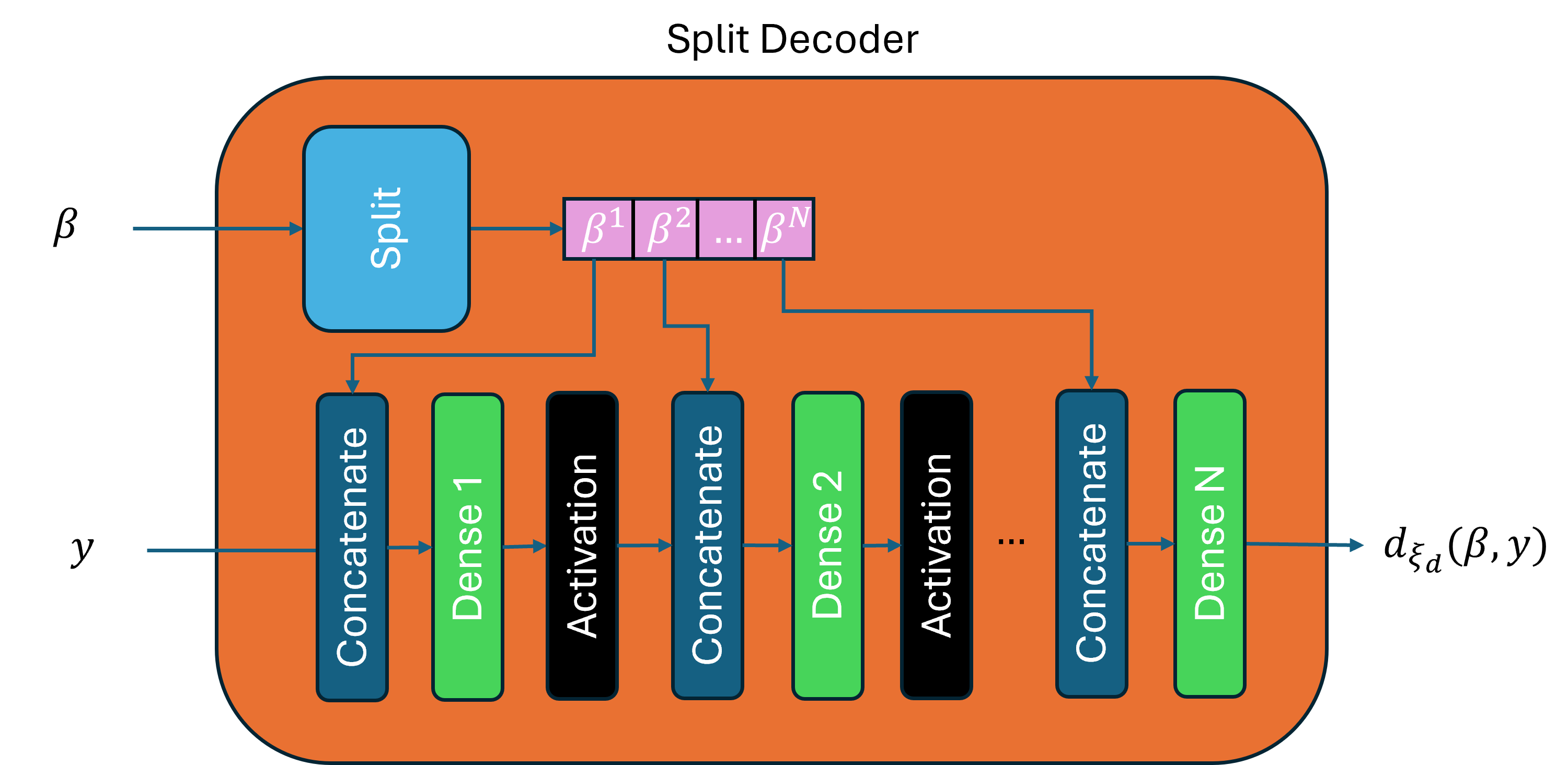}
    \caption{Diagrams representing the two decoders used in the experiments considered in this paper. On the left, the Concat Decoder concatenates $\beta$ and $y$, feeding this larger vector into an MLP. On the right, the Split Decoder breaks $\beta$ into smaller components $\beta^1,\dots, \beta^N$ which are progressively concatenated and fed into intermediate layers of an MLP.}
    \label{fig:decoders}
\end{figure}

The basic formulation of NEON entails base networks that follows an encoder/decoder structure $\mu_\xi(u, y) = d_{\xi_d}(\beta, y) = d_{\xi_d}(e_{\xi_e}(u), y)$, where $\xi = (\xi_e, \xi_d)$ are the trainable parameters of the encoder and decoder, respectively. Although NEON allows for many different choices of architecture for the encoder and decoder, in the experiments carried out in this paper encoders are always MLPs, while decoders are either Concat Decoders or Split Decoders, which are described in this section and illustrated in Figure \ref{fig:decoders}. Other reasonable choices would be convolutional network or a Fourier Neural Operator (FNO)\cite{li2021_FNO} for the encoder, and attention-based decoders\cite{kissas2022_LOCA,rebain2022attention}.

The simpler among the two decoders we used, Concat Decoder is defined by the concatenation of the latent representation $\beta\in\R^{d_\beta}$ of the input function $u$ with a Fourier feature encoding of the query point $y\in \mathcal{Y}$. Although this architecture yields good results on simpler tasks, it may under-perform in cases where a larger latent dimension $d_\beta$ is needed \cite{rebain2022attention}. This would be the case when the target output function concentrate on a high-dimensional nonlinear manifold \cite{seidman2022nomad}.

The Split Decoder approach, inspired by \cite{rebain2022attention}, instead splits $\beta$ into smaller components $\beta=(\beta^1, \beta^2, \dots, \beta^N)$ where $\beta^i\in\R^{d_\beta / N}$. These $\beta^i$ are then used to modulate the features of each hidden layer in the decoder. This approach helped the authors better fit the data by having the hyperparameter $d_\beta$ be larger than it otherwise would be able to. This larger value of $d_\beta$ allowed the encoder to preserve more information about the input function $u$ and create a richer latent representation.

\subsectionPlus{Bayesian Optimization Acquisition Functions}
In this section we examine some possible choices for the acquisition function $\alpha$ used in BO. As previously described, BO aims to optimize a black-box function $f:X\to\R$. Given a trained epistemic model $G_\theta:X\times \mathcal{Z}\to \R$, different choices of acquisition function $\alpha:X\to\R$ can be made. In the case of Operator Composite BO, we have that $G_\theta(u,z)=g\left(\hat{h}_\theta(u,\cdot,z)\right)$, where $g:C(\mathcal{Y},\R^{d_s})\to\R$ is a known and cheap-to-evaluate functional and $\hat{h}_\theta:X\times \mathcal{Y}\times \mathcal{Z}\to \R^{d_s}$ is a neural epistemic operator network trained on the available data.

Popular choices in the literature are Expected Improvement (EI), Probability of Improvement (PI) and Lower Confidence Bound (LCB), also called Upper Confidence Bound (UCB) in the case of function minimization instead of maximization. Some of these common choices are summarized in Table \ref{tab:acquisition_functions}. It should be noted that the definition of these functions is the same in both traditional and composite BO.

An important property of all the acquisition functions considered in this paper and detailed in Table \ref{tab:acquisition_functions} is that they are expectations over $z$. That is, they can be written as $\alpha(u)=\mathbb{E}_{z\sim P_Z}[\alpha'(u,z)]$, where $\alpha':X\times\mathcal{Z}\to\R$. Therefore, we can approximate their value by averaging the quantity of interest over different epistemic indices $z_1,\dots, z_k$ sampled in an \textit{i.i.d.} manner from $P_z$. This is important, as the predictive distribution generated by $G_\theta$ is often intractable or non-analytical. Using a Monte Carlo approximation, we can obtain $\alpha(u) \approx \frac{1}{k}\sum_{i=1}^k \alpha'(u,z_i)$.

The EI acquisition function focuses on \textit{improvement} over the best value acquired so far. That is, if $\{(u_i,f(u_i))\}_{i=1}^N$ is the data collected so far and we set $y_*:=\max\limits_{i=1,\dots,N} f(u_i)$, then EI is defined in terms of our belief of surpassing this value. As the name suggests, we compute the expected improvement from collecting this new point $\alpha_{EI}(u)=\mathbb{E}[\max\{0, f(u)-y_*\}]=\mathbb{E}_{z\sim P_Z}[\max\{0, G_\theta(u,z)-y_*\}]$, where this is determined by the surrogate model $G_\theta$. This can be equivalently expressed as $\alpha_{EI}(u)=\mathbb{E}_{z\sim P_Z}[\texttt{ReLU}(G_\theta(u,z)-y_*)]$, where $\texttt{ReLU}(x):=\max\{0,x\}$ is the Rectified Linear Unit function.

It is well known that functions involving \texttt{ReLU} often suffer from \textit{vanishing gradients}\cite{2012vanishinggradient}, where its derivative is 0 in large regions of the input space. This phenomena is problematic when carrying out optimization with algorithms that use information about derivatives of a function. In particular, we've found that this completely prevented any meaningful optimization of the EI acquisition function in our experiments. In light of this fact, in this paper we also propose the Leaky Expected Improvement (L-EI) acquisition function, where \texttt{ReLU} is substituted by the \texttt{LeakyReLU} \cite{maas2013leakyrelu} in the formulation of $\alpha$. This greatly facilitates the process of determining $u_{\text{new}}=\argmax_{u\in X}\alpha_{\text{L-EI}}(u)$, while maintaining similar global optima as $\alpha_{EI}$. Thus, we obtain $\alpha_{\text{L-EI}}(u)=\mathbb{E}[\alpha_{\text{L-EI}}'(u,z)]$ where
\begin{equation}
    \alpha_{\text{L-EI}}'(u,z) = \texttt{LeakyReLU}(G_\theta(u,z)-y_*) = \begin{cases}
        G_\theta(u,z)-y_* & \text{if }G_\theta(u,z)-y_* \geq 0\\
        0.01(G_\theta(u,z)-y_*) & \text{if }G_\theta(u,z)-y_* < 0
    \end{cases}
\end{equation}
Since the number $0.01$ in the formulation of $\alpha_{\text{L-EI}}'$ is arbitrary, this can be generalized to
\begin{equation}
    \alpha_{\text{L-EI}}'^\delta(u,z) = \texttt{LeakyReLU}^\delta(G_\theta(u,z)-y_*) = \begin{cases}
        G_\theta(u,z)-y_* & \text{if }G_\theta(u,z)-y_* \geq 0\\
        \delta(G_\theta(u,z)-y_*) & \text{if }G_\theta(u,z)-y_* < 0
    \end{cases}
\end{equation}
where we substitute $0.01$ with an arbitrary $\delta>0$. In cases where the subscript $\delta$ is omitted, we assume the default choice of $\delta=0.01$. This generalization allows us to adjust the negative slope of this acquisition function and make it closer to the original formulation of expected improvement if desired. We prove in the appendix the following short theorem, which states that for any given bounded function $f$ we can make Leaky Expected Improvement as close to EI as desired, no matter the choice of surrogate model $G_\theta:X\times\mathcal{Z}\to \R$.

\begin{theorem}\label{thm:l-ei}
    Let $\epsilon>0$ and $f:X\to\R$ be a bounded function. Then there exists a choice of $\delta>0$ such that for any surrogate model $G_{\theta}$ we have that $|\alpha'^{\delta}_\text{L-EI}(x,z) - \alpha'_\text{EI}(x,z)|<\epsilon$. This implies that $|\alpha^{\delta}_\text{L-EI}(x) - \alpha_\text{EI}(x)|<\epsilon$.
\end{theorem}

To the best of the authors' knowledge, this Leaky Expected Improvement (L-EI) acquisition function has not been presented before, but is similar in flavor to the techniques presented in \cite{ament2024unexpected}.

The LCB acquisition function differs from EI by having a hyperparameter $\beta>0$ which explicitly controls the exploration/exploitation trade-off. The LCB acquisition function is composed of two terms: $\mu(u)=\mathbb{E}_{u\sim P_Z}[G_\theta(u,z)]$, which represents the expected value of $f(u)$ according to our surrogate model, and $\sigma(u)=\sqrt{Var_{z\sim P_Z}[G_\theta(u,z)]}$, which represents the epistemic uncertainty of the prediction. It is also common to instead set $\sigma(u)=\mathbb{E}_{u\sim P_Z}\left[\left|G_\theta(u,z)-\mu(u)\right|\right]$. By combining these two terms under a scaling controlled by $\beta$, we obtain $\alpha_{LCB}(u):=\mu(u)+\beta\sigma(u)$, where a higher choice of $\beta$ leads to prioritizing regions of space where our predictions are most uncertain, and lower choices of $\beta$ lead to ignoring uncertainty and focusing on regions that have the highest mean prediction. The value of $\beta$ may be fixed for the entire BO process, or dependent on the iteration number $t$. In particular, scaling $\beta_t\propto \log(t^2)$ leads to provable guarantees for the regret in the GP bandit setting\cite{gp_bandits_2009}.

\begin{table}[ht]
\centering
\begin{tabular}{|c|c|c|}
\hline
Acquisition Function $\alpha$ & Hyperparameters & ENN Formulation\\
\hline
Expected Improvement (EI) & None & $\mathbb{E}_{z\sim P_Z}[\texttt{ReLU}\left(G_\theta(x,z)-y_*\right)]$\\
\hline
Leaky Expected Improvement (L-EI) & $\delta  \in \R_{>0}$ & $\mathbb{E}_{z\sim P_Z}[\texttt{LeakyReLU}^\delta\left(G_\theta(x,z)-y_*\right)]$\\
\hline
Lower Confidence Bound (LCB) & $\beta \in \R_{>0}$ & $\mathbb{E}_{z\sim P_Z}[G_\theta(x,z)] + \beta\sqrt{Var_{z\sim P_Z}[G_\theta(x,z)]}$\\
\hline
\end{tabular}
\caption{\label{tab:acquisition_functions}Possible choices of acquisition functions for Bayesian Optimization. In the case of of EI and L-EI, $y_*=\max\limits_{i=1,\dots,N} f(u_i)$ is the best objective value found so far. In the case of LCB, $\beta>0$ is a hyperparameter that controls the exploration/exploitation trade-off.}
\end{table}

\subsubsectionPlus{Optimization of Acquisition Functions}

In order to carry out BO, it is necessary to compute $u_{\text{new}}:=\argmax_{u\in X}\alpha(u)$ at each iteration, which entails solving an optimization problem. As described in the previous section, values of $\alpha(u)$ are determined in a Monte Carlo fashion as 
\begin{equation}
\alpha(u) = \mathbb{E}_{z\sim P_Z}[\alpha'(u,z)] \approx \frac{1}{k}\sum_{i=1}^k \alpha'(u,z_i), \ \ z_i\overset{\text{\textit{iid}}}{\sim} P_Z.
\end{equation}
Since all models and acquisition functions considered in this paper are almost-everywhere differentiable as a function of $u\in X$, we optimize $\alpha$ using the SciPy\cite{2020SciPy-NMeth} implementation of the constrained Limited memory BFGS (L-BFGS) algorithm\cite{liu1989lbfgs}. We carry out this process $n_\text{reset}$ times using different initial points $u_0^{1},\dots, u_0^{n_\text{reset}}$ to obtain converged solutions $u_{\text{final}}^1, \dots, u_{\text{final}}^{n_\text{reset}}$ and set the new point to be collected as $u_{\text{new}} = \argmax_{i=1,\dots, n_\text{reset}}\alpha(u_{\text{final}}^i)$. Due to the highly non-convex nature of the problems considered in this paper, $n_\text{reset}$ is an important hyperparameter to consider, and we've found that setting $n_\text{reset}=500$ yields good results in practice. It should also be noted that each optimization $u_0^i\mapsto u_\text{final}^i$ can be done independently, so the process of obtaining all the $x_\text{final}^i$ is easily parallelised.

\subsubsectionPlus{Parallel Acquisition}

The acquisition functions described in Table \ref{tab:acquisition_functions} assume that $q=1$ new points are acquired at each iteration of BO. Extensions of these methods such as $q$-EI\cite{qEI} and more\cite{daulton2020differentiable, wang2016parallel} propose to determine $\argmax_{x_1,\dots,x_q\in X^q}\alpha_q(x_1,\dots,x_q)$ where at each step $q\geq 1$ different inputs are acquired simultaneously. These multi-acquisition functions are also compatible with the NEON framework and an experiment in this setting is briefly explored in the appendix.

\sectionPlus{Results}

In this section we employ our method on several synthetic and realistic benchmarks. By augmenting a deterministic operator network with a small EpiNet, thus creating NEON, we observe considerably less data and model-size requirements in order to obtain good performance. Across all experiments we observe comparable or better performance to state-of-the-art approaches, while using orders of magnitude less trainable parameters. A comparison on the total number of trainable parameter can be seen in Table \ref{tab:parameters}.

In what follows, we wish to optimize a function $f:X\to \R$ with compositional structure $f=g\circ h$ where $X\subseteq \R^{d_u}$ is a finite dimensional space, $h:X\to C(\mathcal{Y}, \R^{d_s})$ is an unknown map from $X$ to a space of continuous functions, and $g:C(\mathcal{Y},\R^{d_s})\to \R$ is a known and cheap-to-evaluate functional. In practice, we must evaluate the function $h(u)$ on several points $y_1,\dots,y_m\in \mathcal{Y}$ in order to compute $f(u)=g(h(u))$. Although NEON allows for the evaluation of $h(u)$ on any arbitrary point $y\in\mathcal{Y}$, we limit ourselves to a fixed grid in order to be compatible with the existing literature\cite{maddox2021bayesian, BHOURI2023116428} and provide a fair comparison. We then wish to optimize $f(u):=g(h(u))$ using the procedure described in the Methods section.

\begin{figure}[ht]
    \centering
    \includegraphics[width=0.4\textwidth]{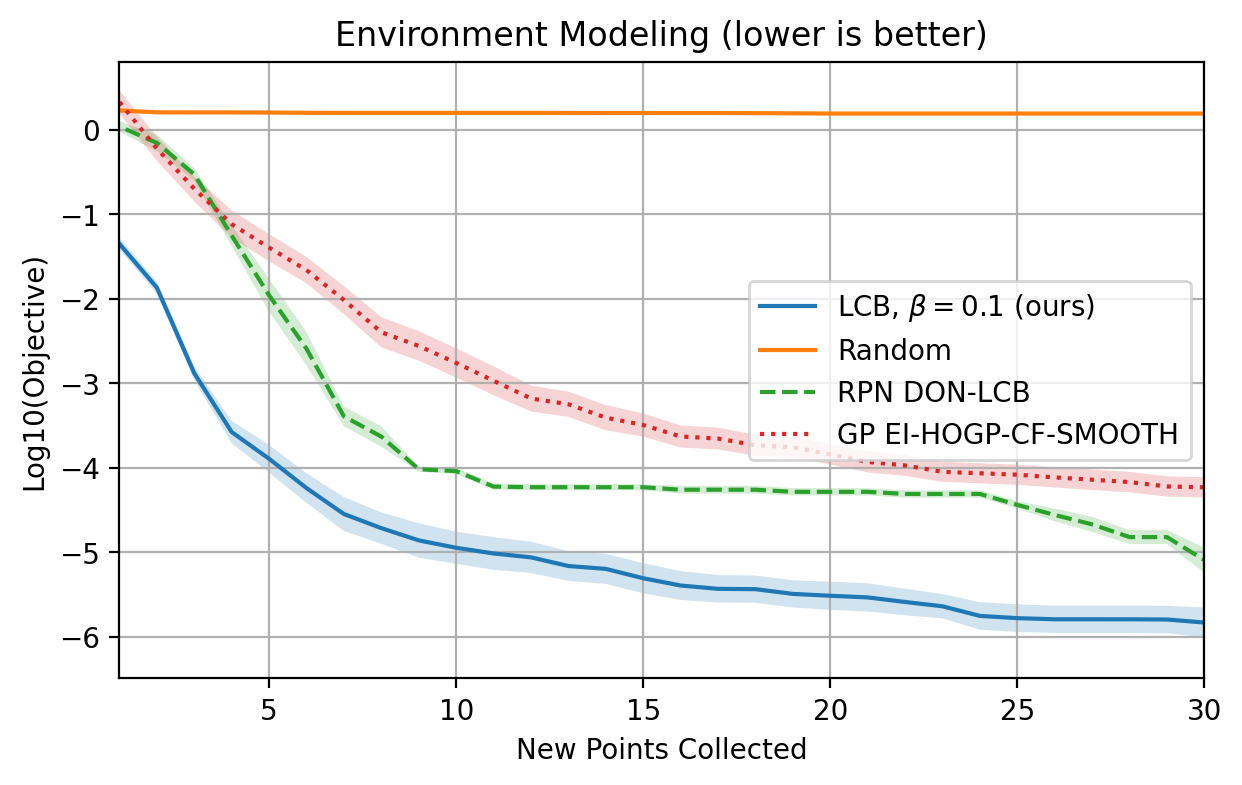}
    \includegraphics[width=0.4\textwidth]{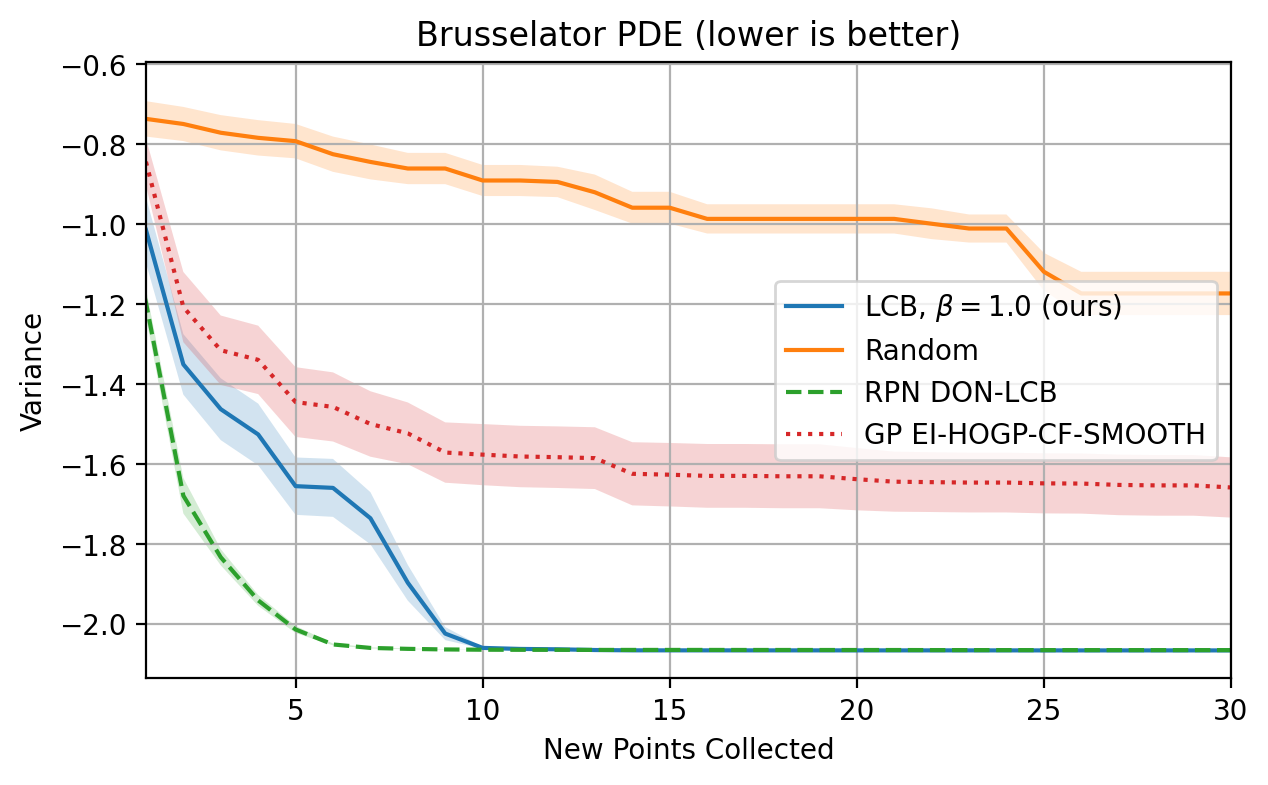}
    \caption{Experimental results for the Environment Model (left) and Brusselator PDE (right) problems. In both cases we plot the acquisition function that resulted in the lowest final mean objective across 10 different seeds for NEON, RPN\cite{BHOURI2023116428} and GP\cite{maddox2021bayesian} approaches. Remaining comparisons across different methods can be seen in the appendix. As can be seen, our approach performs significantly better on the environmental model problem, and comparably well on the Brusselator PDE. Following \cite{BHOURI2023116428}, the uncertainty bands indicate 20\% of the standard deviation band.}
    \label{fig:results_pollutants_brusselator}
\end{figure}

\subsectionPlus{Environment Model Function}

This example consists of modeling a chemical spill in a long and narrow river, and is a common synthetic benchmark in the Bayesian optimization literature\cite{astudillo2019bayesian, BHOURI2023116428}. Given the concentration of a chemical along the river at different points in time and space, the objective of this problem is to determine the original conditions of the spills: total mass, time, position and diffusion rate. For this problem, $d_u=4$ with $X=\subset \R^{d_u}$  and $h:X\to C(\mathcal{Y},\R)$, with $h(u)$ being evaluated on a 3x4 grid on the domain $\mathcal{Y}=[0,2.5]\times[15,60] \subset \R^2$. Further details about the underlying process can be found in the appendix as well as in \cite{baysean-calibration-2008, astudillo2019bayesian}.

\subsectionPlus{Brusselator PDE}

This problem attempts to determine the optimal diffusion and reaction rate parameters to stabilize a chemical dynamical system. This dynamical system can be modeled according to a PDE described in the \textit{py-pde}\cite{py-pde} documentation \href{https://py-pde.readthedocs.io/en/latest/examples_gallery/pde_brusselator_expression.html}{https://py-pde.readthedocs.io/en/latest/examples\_gallery/pde\_brusselator\_expression.html}. The goal is then to minimize the the weighted variance of of the solution of this PDE, which can be thought of as finding a stable configuration of the chemical reaction. For this problem, $d_u=4$ with $X=[0.1,5]^2\times[0.01,5]^2\subset \R^{d_u}$  and $h:X\to C(\mathcal{Y},\R^2)$, with $h(u)$ being evaluated on a 64x64 grid on the domain $\mathcal{Y}=[0,1]^2\subset\R^2$. Ground truth solutions are obtained via the \textit{py-pde}\cite{py-pde} package.

\begin{figure}[ht]
    \centering
    \includegraphics[width=0.4\textwidth]{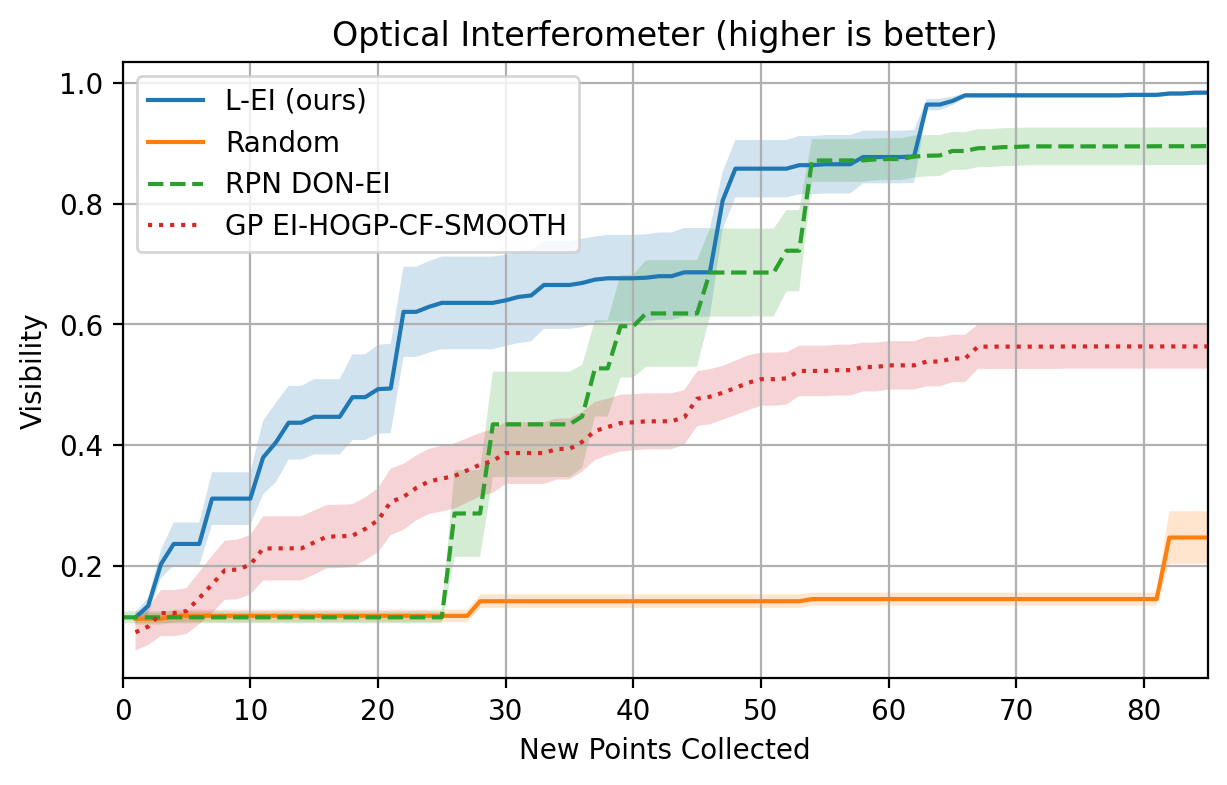}
    \includegraphics[width=0.4\textwidth]{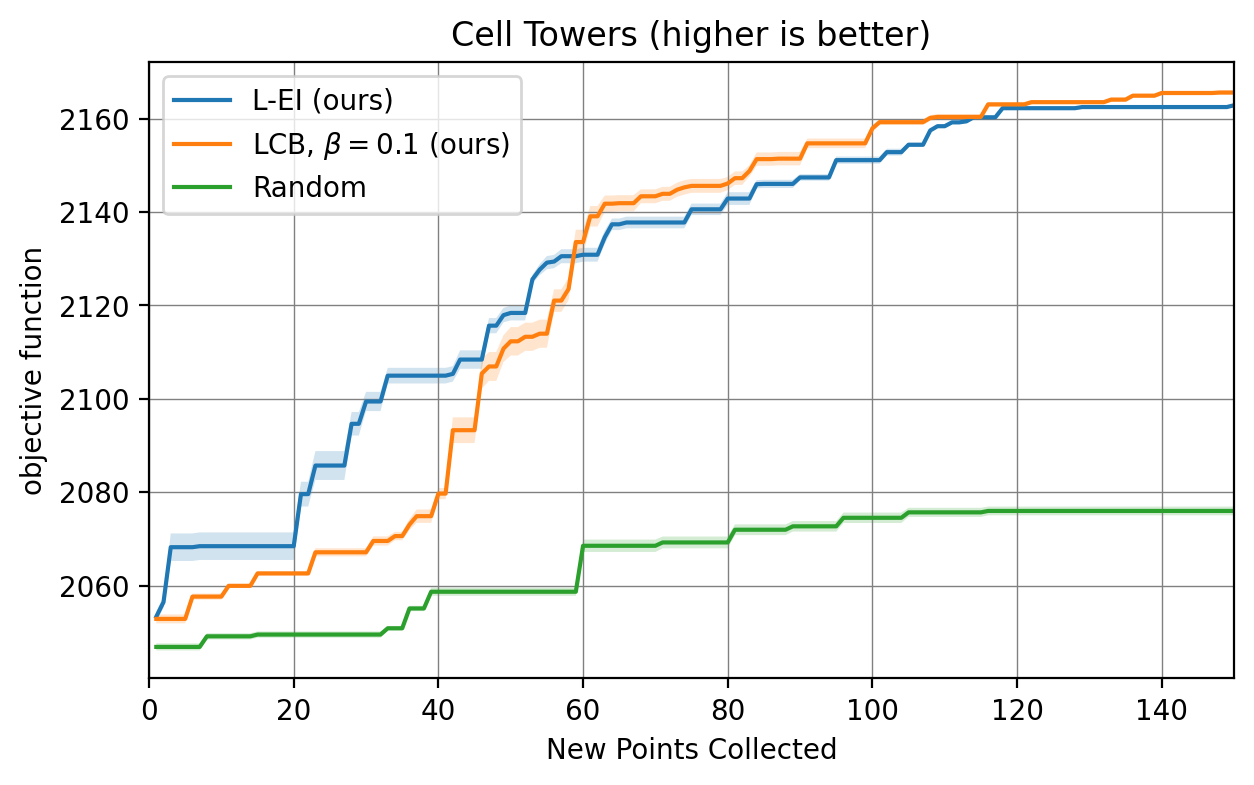}
    \caption{Experimental results for the Optical Interferometer (left) and Cell Towers (right) problems. For the interferometer case, we plot the acquisition function that resulted in the lowest final mean objective across 5 different seeds for NEON, RPN\cite{BHOURI2023116428} and GP\cite{maddox2021bayesian} approaches. For the cell towers case, we plot comparisons between the L-EI acquisition function and LCB using $\beta=0.1$. Remaining comparisons across different methods can be seen in the appendix. Our approach performs significantly better on the optical interferometer problem, and indicate good behaviour on the cell towers problem. Following \cite{BHOURI2023116428}, the uncertainty bands indicate 20\% of the standard deviation band.}
    \label{fig:results_interferometer_cell}
\end{figure}

\subsectionPlus{Optical Interferometer}

This problem consist of aligning a  Mach-Zehnder interferometer by determining the positional parameters of two mirrors in order to minimize optical interference and thus increase visibility, which is measured as defined in \cite{sorokin2021interferobot}. For this problem, $d_u=4$ with $X=[-1,1]^4\subset \R^{d_u}$  and $h:X\to C(\mathcal{Y},\R^{16})$, with $h(u)$ being evaluated on a 64x64 grid on the domain $\mathcal{Y}=[0,1]^2\subset\R^2$. Ground truth solutions are obtained using the code from \cite{sorokin2021interferobot}, available at \href{https://github.com/dmitrySorokin/interferobotProject}{https://github.com/dmitrySorokin/interferobotProject}.

\subsectionPlus{Cell Towers}

This problem consists of optimizing parameters of 15 different cell service towers in order to maximize coverage area and minimize signal interference between towers. Each tower has a downtilt angle parameter lying in $(0,10)$, as well as a power level parameter lying in $(30,50)$, meaning that the input dimension of this problem is $d_u=30$, with inputs $u\in (0,10)^{15}\times (30,50)^{15}$. The function $h:X \to C(\R^2,\R^2)$ then uses these parameters to compute the map of coverage and interference strengths, respectively. An example of such maps can be seen in Figure \ref{fig:cell_plots}. We compute the ground truth values of $h(u)$ using the code presented in \cite{dreifuerst2021optimizing} and available at \href{https://github.com/Ryandry1st/CCO-in-ORAN}{https://github.com/Ryandry1st/CCO-in-ORAN}, which creates results in a $241\times 241$ grid, then downsampling to $50\times 50$, as done in \cite{maddox2021bayesian}. We then use the objective function $g:C(\R^2,\R^2)\to\R$ presented in \cite{maddox2021bayesian} to obtain the final score $f(u)=g(h(u))\in \R$ for a given set of parameters $u$.

\sectionPlus{Discussion}

%\textbf{[The Discussion should be succinct and must not contain subheadings.]}

In this paper we propose a unified framework for understanding uncertainty in operator learning and high-dimensional regression, based on the general language of Epistemic Neural Networks (ENNs). We combine ideas from ENNs and operator learning to create NEON, a general framework for epistemic uncertainty quantification in neural operators using a single backbone architecture, as opposed to an ensemble as is done in \cite{deep-ensambles-2016, BHOURI2023116428}. NEON works by augmenting any operator network with an EpiNet, a small neural network as described in \cite{epistemicNNs}, which allows for easy and inexpensive quantification of epistemic uncertainty.

By carrying out numerical experiments on four different benchmarks for operator composite BO, we demonstrate NEON achieves similar or better performance to state-of-the-art methods while often using orders of magnitude less training parameters, as can be seen in Table \ref{tab:parameters}.

In the more complex case of the Optical Interferometer, an ensemble of small networks is generally unable to capture the complex nature of the map $h$. As such, we use a single larger network augmented by a small EpiNet (195,728 trainable parameters in total) instead of an ensemble of small networks as was done in \cite{BHOURI2023116428}. This led to the biggest improvement in performance between the two methods among the cases considered. The implementation of NEON allows for the added power of a larger network while still capturing epistemic uncertainty, whereas an ensemble of equally powerful networks would present a much higher computational cost.

\begin{table}[ht]
\centering
\begin{tabular}{|c|c|c|c|c|}
\hline
Model$\backslash$Problem & Pollutants & PDE Brusselator & Optical Interferometer & Cell Towers\\
\hline
NEON (ours) & 27,746 & 38,820 & 195,728 & 93,096\\
\hline
RPN DeepONet Ensamble \cite{BHOURI2023116428} & 1,123,072 & 1,131,264 & 45,152 & N/A\\
\hline
\end{tabular}
\caption{\label{tab:parameters}Number of trainable parameters for models used in this paper and in \cite{BHOURI2023116428}.}
\end{table}

Future research in this area may focus on finding more appropriate architectures for EpiNets carrying out UQ in high dimensions, or studying new applications to this method, such as active learning for operator networks.

\section*{Acknowledgements}

%Acknowledgements should be brief, and should not include thanks to anonymous referees and editors, or effusive comments. Grant or contribution numbers may be acknowledged.
We would like to acknowledge support from the US Department of Energy under the Advanced Scientific Computing Research program (grant DE-SC0024563), the US National Science Foundation (NSF) Soft AE Research Traineeship (NRT) Program (NSF grant 2152205) and the Samsung GRO Fellowship program. We also thank Shyam Sankaran for helpful feedback when reviewing the manuscript and the developers of software that enabled this research, including JAX\cite{jax2018github}, Flax\cite{flax2020github} Matplotlib\cite{Hunter2007matplotlib} and NumPy\cite{harris2020numpy}.

%Must include all authors, identified by initials, for example:
%A.A. conceived the experiment(s),  A.A. and B.A. conducted the experiment(s), C.A. and D.A. analysed the results.  All authors reviewed the manuscript. 

%\section*{Additional information}

%To include, in this order: \textbf{Accession codes} (where applicable); \textbf{Competing interests} (mandatory statement). 

%The corresponding author is responsible for submitting a \href{http://www.nature.com/srep/policies/index.html#competing}{competing interests statement} on behalf of all authors of the paper. This statement must be included in the submitted article file.

%Bibliography
\bibliographystyle{unsrt}  
\bibliography{bib}

\begin{thebibliography}{10}

\bibitem{RasmussenW06_gps4ml}
Carl~Edward Rasmussen and Christopher K.~I. Williams.
\newblock {\em Gaussian processes for machine learning.}
\newblock Adaptive computation and machine learning. MIT Press, 2006.

\bibitem{bnn-survey-2020}
Ethan Goan and Clinton Fookes.
\newblock Bayesian neural networks: An introduction and survey.
\newblock In {\em Case Studies in Applied Bayesian Data Science}, pages 45--87. Springer International Publishing, 2020.

\bibitem{deep-ensambles-2016}
Balaji Lakshminarayanan, Alexander Pritzel, and Charles Blundell.
\newblock Simple and scalable predictive uncertainty estimation using deep ensembles, 2016.

\bibitem{Lu_2021_deeponet}
Lu~Lu, Pengzhan Jin, Guofei Pang, Zhongqiang Zhang, and George~Em Karniadakis.
\newblock Learning nonlinear operators via {DeepONet} based on the universal approximation theorem of operators.
\newblock {\em Nature Machine Intelligence}, 3(3):218--229, mar 2021.

\bibitem{li2021_FNO}
Zongyi Li, Nikola Kovachki, Kamyar Azizzadenesheli, Burigede Liu, Kaushik Bhattacharya, Andrew Stuart, and Anima Anandkumar.
\newblock Fourier neural operator for parametric partial differential equations, 2021.

\bibitem{kovachki2023neural_operator}
Nikola Kovachki, Zongyi Li, Burigede Liu, Kamyar Azizzadenesheli, Kaushik Bhattacharya, Andrew Stuart, and Anima Anandkumar.
\newblock Neural operator: Learning maps between function spaces with applications to pdes.
\newblock {\em Journal of Machine Learning Research}, 24(89):1--97, 2023.

\bibitem{kissas2022_LOCA}
Georgios Kissas, Jacob~H Seidman, Leonardo~Ferreira Guilhoto, Victor~M Preciado, George~J Pappas, and Paris Perdikaris.
\newblock Learning operators with coupled attention.
\newblock {\em The Journal of Machine Learning Research}, 23(1):9636--9698, 2022.

\bibitem{wang2021learning}
Sifan Wang, Hanwen Wang, and Paris Perdikaris.
\newblock Learning the solution operator of parametric partial differential equations with physics-informed deeponets.
\newblock {\em Science advances}, 7(40):eabi8605, 2021.

\bibitem{wang2022improved}
Sifan Wang, Hanwen Wang, and Paris Perdikaris.
\newblock Improved architectures and training algorithms for deep operator networks.
\newblock {\em Journal of Scientific Computing}, 92(2):35, 2022.

\bibitem{yang2022scalable}
Yibo Yang, Georgios Kissas, and Paris Perdikaris.
\newblock Scalable uncertainty quantification for deep operator networks using randomized priors.
\newblock {\em Computer Methods in Applied Mechanics and Engineering}, 399:115399, 2022.

\bibitem{psaros2023uncertainty}
Apostolos~F Psaros, Xuhui Meng, Zongren Zou, Ling Guo, and George~Em Karniadakis.
\newblock Uncertainty quantification in scientific machine learning: Methods, metrics, and comparisons.
\newblock {\em Journal of Computational Physics}, 477:111902, 2023.

\bibitem{filos2019systematic}
Angelos Filos, Sebastian Farquhar, Aidan~N. Gomez, Tim G.~J. Rudner, Zachary Kenton, Lewis Smith, Milad Alizadeh, Arnoud de~Kroon, and Yarin Gal.
\newblock A systematic comparison of bayesian deep learning robustness in diabetic retinopathy tasks, 2019.

\bibitem{Esteva2017dermatologist}
Andre Esteva, Brett Kuprel, Roberto~A. Novoa, Justin Ko, Susan~M. Swetter, Helen~M. Blau, and Sebastian Thrun.
\newblock Dermatologist-level classification of skin cancer with deep neural networks.
\newblock {\em Nature}, 542(7639):115–118, January 2017.

\bibitem{huang2020autonomous}
Yu~Huang and Yue Chen.
\newblock Autonomous driving with deep learning: A survey of state-of-art technologies, 2020.

\bibitem{BALD-2011}
Neil Houlsby, Ferenc Huszár, Zoubin Ghahramani, and Máté Lengyel.
\newblock Bayesian active learning for classification and preference learning, 2011.

\bibitem{batchbald-2019}
Andreas Kirsch, Joost van Amersfoort, and Yarin Gal.
\newblock Batchbald: Efficient and diverse batch acquisition for deep bayesian active learning, 2019.

\bibitem{epistemicNNs}
Ian Osband, Zheng Wen, Mohammad Asghari, Morteza Ibrahimi, Xiyuan Lu, and Benjamin~Van Roy.
\newblock Epistemic neural networks.
\newblock {\em CoRR}, abs/2107.08924, 2021.

\bibitem{wang2022boreview}
Xilu Wang, Yaochu Jin, Sebastian Schmitt, and Markus Olhofer.
\newblock Recent advances in bayesian optimization, 2022.

\bibitem{botorch}
Maximilian Balandat, Brian Karrer, Daniel~R. Jiang, Samuel Daulton, Benjamin Letham, Andrew~Gordon Wilson, and Eytan Bakshy.
\newblock Botorch: Programmable bayesian optimization in pytorch.
\newblock {\em CoRR}, abs/1910.06403, 2019.

\bibitem{gal2016dropout}
Yarin Gal and Zoubin Ghahramani.
\newblock Dropout as a bayesian approximation: Representing model uncertainty in deep learning, 2016.

\bibitem{seidman2022nomad}
Jacob~H. Seidman, Georgios Kissas, Paris Perdikaris, and George~J. Pappas.
\newblock Nomad: Nonlinear manifold decoders for operator learning, 2022.

\bibitem{BHOURI2023116428}
Mohamed~Aziz Bhouri, Michael Joly, Robert Yu, Soumalya Sarkar, and Paris Perdikaris.
\newblock Scalable bayesian optimization with randomized prior networks.
\newblock {\em Computer Methods in Applied Mechanics and Engineering}, 417:116428, 2023.

\bibitem{maddox2021bayesian}
Wesley~J Maddox, Maximilian Balandat, Andrew~G Wilson, and Eytan Bakshy.
\newblock Bayesian optimization with high-dimensional outputs.
\newblock {\em Advances in neural information processing systems}, 34:19274--19287, 2021.

\bibitem{chenchen1995}
Tianping Chen and Hong Chen.
\newblock Universal approximation to nonlinear operators by neural networks with arbitrary activation functions and its application to dynamical systems.
\newblock {\em IEEE Transactions on Neural Networks}, 6(4):911--917, 1995.

\bibitem{di2023neural}
Patricio~Clark Di~Leoni, Lu~Lu, Charles Meneveau, George~Em Karniadakis, and Tamer~A Zaki.
\newblock Neural operator prediction of linear instability waves in high-speed boundary layers.
\newblock {\em Journal of Computational Physics}, 474:111793, 2023.

\bibitem{jin2022mionet}
Pengzhan Jin, Shuai Meng, and Lu~Lu.
\newblock Mionet: Learning multiple-input operators via tensor product, 2022.

\bibitem{astudillo2019bayesian}
Raul Astudillo and Peter~I. Frazier.
\newblock Bayesian optimization of composite functions, 2019.

\bibitem{maus2023joint}
Natalie Maus, Zhiyuan~Jerry Lin, Maximilian Balandat, and Eytan Bakshy.
\newblock Joint composite latent space bayesian optimization.
\newblock {\em arXiv preprint arXiv:2311.02213}, 2023.

\bibitem{dreifuerst2021optimizing}
Ryan~M Dreifuerst, Samuel Daulton, Yuchen Qian, Paul Varkey, Maximilian Balandat, Sanjay Kasturia, Anoop Tomar, Ali Yazdan, Vish Ponnampalam, and Robert~W Heath.
\newblock Optimizing coverage and capacity in cellular networks using machine learning.
\newblock In {\em ICASSP 2021-2021 IEEE International Conference on Acoustics, Speech and Signal Processing (ICASSP)}, pages 8138--8142. IEEE, 2021.

\bibitem{kim2022deep}
Samuel Kim, Peter~Y Lu, Charlotte Loh, Jamie Smith, Jasper Snoek, and Marin Solja{\v{c}}i{\'c}.
\newblock Deep learning for bayesian optimization of scientific problems with high-dimensional structure.
\newblock {\em Transactions on Machine Learning Research}, 2022.

\bibitem{tancik2020fourfeat}
Matthew Tancik, Pratul~P. Srinivasan, Ben Mildenhall, Sara Fridovich-Keil, Nithin Raghavan, Utkarsh Singhal, Ravi Ramamoorthi, Jonathan~T. Barron, and Ren Ng.
\newblock Fourier features let networks learn high frequency functions in low dimensional domains.
\newblock {\em NeurIPS}, 2020.

\bibitem{rebain2022attention}
Daniel Rebain, Mark~J. Matthews, Kwang~Moo Yi, Gopal Sharma, Dmitry Lagun, and Andrea Tagliasacchi.
\newblock Attention beats concatenation for conditioning neural fields, 2022.

\bibitem{2012vanishinggradient}
Razvan Pascanu, Tom{\'{a}}s Mikolov, and Yoshua Bengio.
\newblock Understanding the exploding gradient problem.
\newblock {\em CoRR}, abs/1211.5063, 2012.

\bibitem{maas2013leakyrelu}
Andrew~L Maas, Awni~Y Hannun, Andrew~Y Ng, et~al.
\newblock Rectifier nonlinearities improve neural network acoustic models.
\newblock In {\em Proc. icml}, volume 30-1, page~3. Atlanta, GA, 2013.

\bibitem{ament2024unexpected}
Sebastian Ament, Samuel Daulton, David Eriksson, Maximilian Balandat, and Eytan Bakshy.
\newblock Unexpected improvements to expected improvement for bayesian optimization.
\newblock {\em Advances in Neural Information Processing Systems}, 36, 2024.

\bibitem{gp_bandits_2009}
Niranjan Srinivas, Andreas Krause, Sham~M. Kakade, and Matthias~W. Seeger.
\newblock Gaussian process bandits without regret: An experimental design approach.
\newblock {\em CoRR}, abs/0912.3995, 2009.

\bibitem{2020SciPy-NMeth}
Pauli Virtanen, Ralf Gommers, Travis~E. Oliphant, Matt Haberland, Tyler Reddy, David Cournapeau, Evgeni Burovski, Pearu Peterson, Warren Weckesser, Jonathan Bright, St{\'e}fan~J. {van der Walt}, Matthew Brett, Joshua Wilson, K.~Jarrod Millman, Nikolay Mayorov, Andrew R.~J. Nelson, Eric Jones, Robert Kern, Eric Larson, C~J Carey, {\.I}lhan Polat, Yu~Feng, Eric~W. Moore, Jake {VanderPlas}, Denis Laxalde, Josef Perktold, Robert Cimrman, Ian Henriksen, E.~A. Quintero, Charles~R. Harris, Anne~M. Archibald, Ant{\^o}nio~H. Ribeiro, Fabian Pedregosa, Paul {van Mulbregt}, and {SciPy 1.0 Contributors}.
\newblock {{SciPy} 1.0: Fundamental Algorithms for Scientific Computing in Python}.
\newblock {\em Nature Methods}, 17:261--272, 2020.

\bibitem{liu1989lbfgs}
Dong~C Liu and Jorge Nocedal.
\newblock On the limited memory bfgs method for large scale optimization.
\newblock {\em Mathematical programming}, 45(1):503--528, 1989.

\bibitem{qEI}
David Ginsbourger, Rodolphe Le~Riche, and Laurent Carraro.
\newblock Kriging is well-suited to parallelize optimization.
\newblock In {\em Computational intelligence in expensive optimization problems}, pages 131--162. Springer, 2010.

\bibitem{daulton2020differentiable}
Samuel Daulton, Maximilian Balandat, and Eytan Bakshy.
\newblock Differentiable expected hypervolume improvement for parallel multi-objective bayesian optimization.
\newblock {\em Advances in Neural Information Processing Systems}, 33:9851--9864, 2020.

\bibitem{wang2016parallel}
Jialei Wang, Scott~C Clark, Eric Liu, and Peter~I Frazier.
\newblock Parallel bayesian global optimization of expensive functions.
\newblock {\em arXiv preprint arXiv:1602.05149}, 2016.

\bibitem{baysean-calibration-2008}
Nikolay Bliznyuk, David Ruppert, Christine Shoemaker, Rommel Regis, Stefan Wild, and Pradeep Mugunthan.
\newblock Bayesian calibration and uncertainty analysis for computationally expensive models using optimization and radial basis function approximation.
\newblock {\em Journal of Computational and Graphical Statistics}, 17(2):270--294, 2008.

\bibitem{py-pde}
David Zwicker.
\newblock py-pde: A python package for solving partial differential equations.
\newblock {\em Journal of Open Source Software}, 5(48):2158, 2020.

\bibitem{sorokin2021interferobot}
Dmitry Sorokin, Alexander Ulanov, Ekaterina Sazhina, and Alexander Lvovsky.
\newblock Interferobot: aligning an optical interferometer by a reinforcement learning agent, 2021.

\bibitem{jax2018github}
James Bradbury, Roy Frostig, Peter Hawkins, Matthew~James Johnson, Chris Leary, Dougal Maclaurin, George Necula, Adam Paszke, Jake Vander{P}las, Skye Wanderman-{M}ilne, and Qiao Zhang.
\newblock {JAX}: composable transformations of {P}ython+{N}um{P}y programs, 2018.

\bibitem{flax2020github}
Jonathan Heek, Anselm Levskaya, Avital Oliver, Marvin Ritter, Bertrand Rondepierre, Andreas Steiner, and Marc van {Z}ee.
\newblock {F}lax: A neural network library and ecosystem for {JAX}, 2023.

\bibitem{Hunter2007matplotlib}
J.~D. Hunter.
\newblock Matplotlib: A 2d graphics environment.
\newblock {\em Computing in Science \& Engineering}, 9(3):90--95, 2007.

\bibitem{harris2020numpy}
Charles~R. Harris, K.~Jarrod Millman, St{\'{e}}fan~J. van~der Walt, Ralf Gommers, Pauli Virtanen, David Cournapeau, Eric Wieser, Julian Taylor, Sebastian Berg, Nathaniel~J. Smith, Robert Kern, Matti Picus, Stephan Hoyer, Marten~H. van Kerkwijk, Matthew Brett, Allan Haldane, Jaime~Fern{\'{a}}ndez del R{\'{i}}o, Mark Wiebe, Pearu Peterson, Pierre G{\'{e}}rard-Marchant, Kevin Sheppard, Tyler Reddy, Warren Weckesser, Hameer Abbasi, Christoph Gohlke, and Travis~E. Oliphant.
\newblock Array programming with {NumPy}.
\newblock {\em Nature}, 585(7825):357--362, September 2020.

\bibitem{bliznyuk2008bayesian}
Nikolay Bliznyuk, David Ruppert, Christine Shoemaker, Rommel Regis, Stefan Wild, and Pradeep Mugunthan.
\newblock Bayesian calibration and uncertainty analysis for computationally expensive models using optimization and radial basis function approximation.
\newblock {\em Journal of Computational and Graphical Statistics}, 17(2):270--294, 2008.

\bibitem{kingma2017adam}
Diederik~P. Kingma and Jimmy Ba.
\newblock Adam: A method for stochastic optimization, 2017.

\bibitem{wilson2018maximizing}
James~T. Wilson, Frank Hutter, and Marc~Peter Deisenroth.
\newblock Maximizing acquisition functions for bayesian optimization, 2018.

\end{thebibliography}

\section*{Author contributions statement}
L.F.G. and P.P. conceived the methodology. L.F.G. conducted the experiments and analysed the results. P.P. provided funding and supervised this study. All authors reviewed the manuscript.

\section*{Competing Interests Statement}

To the best of the author's knowledge, they have no competing interests related to this research.

\section*{Data Availability Statement}

At the time of this preprint, the data and code used in this paper is available upon request to the authors.

\sectionPlus{Proof of Theorem 1}

Since $f:X\to \R$ is bounded, there exists $a<b$ such that $f(x)\in [a,b]$ for all $x\in X$. If we now define $M:=b-a>0$ and set any $\delta < \frac{\epsilon}{M}$ we have that 
\begin{align*}
    \alpha'^\delta_\text{L-EI}(u,z) - \alpha'_\text{EI}(u,z) &=  \begin{cases}
        0, & \text{if }G_\theta(u,z)-y_* \geq 0,\\
        \delta(G_\theta(u,z)-y_*), & \text{if }G_\theta(u,z)-y_* < 0.
    \end{cases}
\end{align*}
Additionally, if we assume that $G_\theta$ is any epistemic model with range within the accepted range $[a,b]$ of $f$, we get that $|G_\theta(u,z)-y_*|\leq M$ and therefore
\begin{align*}
    |\alpha'^\delta_\text{L-EI}(u,z) - \alpha'_\text{EI}(u,z)| &\leq  \delta|G_\theta(u,z)-y_*|\\
        &\leq \delta  M\\
        &< \frac{\epsilon}{M}\cdot M\\
        &= \epsilon,
\end{align*}
which proves the first statement of the theorem. Finally, we have that
\begin{align*}
    |\alpha^\delta_\text{L-EI}(u) - \alpha_\text{EI}(u)| &= \left| \mathbb{E}_z\left[\alpha'^\delta_\text{L-EI}(u,z)\right] - \mathbb{E}_z\left[\alpha'_\text{EI}(u,z)\right]  \right|\\
        &= \left| \mathbb{E}_z\left[\alpha'^\delta_\text{L-EI}(u,z) - \alpha'_\text{EI}(u,z)\right]  \right|\\
        &\leq  \mathbb{E}_z\left[ \left| \alpha'^\delta_\text{L-EI}(u,z) - \alpha'_\text{EI}(u,z)  \right| \right]\\
        &<  \mathbb{E}_z\left[ \epsilon \right]\\
        &= \epsilon,
\end{align*}
which concludes the proof. \hfill$\square$

\sectionPlus{Further Experimental Details}

\subsectionPlus{Environmental Model Function}

As mentioned in the main text, this is an inverse problem which models two chemical pollutant spills in a long and narrow river. Given observations of the chemical concentration at certain points in time and space, the objective of this problem is to determine the initial conditions of the spill. This is a common toy benchmark for composite BO\cite{astudillo2019bayesian, maddox2021bayesian, BHOURI2023116428} following the formulation described in \cite{bliznyuk2008bayesian}.

The true values for the map $h:X\to C(\R^2,\R)$ are given by
\begin{align*}
    h(M,D,L,\tau)(s, t) = \begin{cases}
        \frac{M}{2\sqrt{\pi Dt}}\exp{\left(-\frac{s^2}{4Dt}\right)}, \quad \text{if }t<\tau,\\
        \frac{M}{2\sqrt{\pi Dt}}\exp{\left(-\frac{s^2}{4Dt}\right)} + \frac{M}{2\sqrt{\pi D(t-\tau)}}\exp{\left(-\frac{(s-L)^2}{4D(t-\tau)}\right)}, \quad \text{if }t\geq\tau,
    \end{cases}
\end{align*}
where $M\in(7,13)$ is the pollutant mass spilled at each location, $D\in(0.02,0.12)$ is the diffusion rate of the chemical in water, $L\in(0.01,3)$ is the location of the second spill relative to the first one and $\tau\in (30.01,30.295)$ is the time of the second spill relative to the first one.

Using the true parameters $u_\text{true}=(10,0.07, 15.05, 30.1525)$, the objective of this problem is to determine this value from observations of the chemical concentrations at positions $s\in\{0,1,2.5\}$ for times $t\in\{15, 30, 45, 60\}$.

The NEON architecture for this experiment used an MLP encoder with 2 hidden layers with hidden dimension 64 and $d_\beta=64$. We used a Split Decoder with 2 layers of hidden dimension 64. The EpiNet architecture we used consisted of a trainable MLP with two hidden layers of dimension 32, and for the prior component an ensemble of 16 MLPs with 2 hidden layers of width 5 each and a scale parameter of 0.75. We trained this network for 12,000 steps using a batch size of 256 and the Adam\cite{kingma2017adam} optimizer and exponential learning rate decay.

\subsectionPlus{Brusselator PDE}

The PDE considered in this experiment describes the chemical reaction over time $t$ of two compounds $u(t)$ and $v(t)$
\begin{align*}
    \partial_t u &= D_0\nabla^2 u + a - (1-b)u + vu^2, \\
    \partial_t v &= D_1\nabla^2 v + bu -vu^2,
\end{align*}
where $D_1,D_2\in(0.01,5)$ are the diffusivity parameters of chemicals $u$ and $v$, respectively, and $a,b\in (0.1,5)$ are their reaction rates.

The tuple $(a,b,D_0,D_1)\in X$ is the quantity we wish to minimize over in order to obtain the smallest value for the objective, which is the weighted variance of the system, determining by evolving the PDE from $t=0$ to $t=20$ using the example implementation of the PyPDE\cite{py-pde} package.

The NEON architecture for this experiment used an MLP encoder with 2 hidden layers with hidden dimension 64 and $d_\beta=96$. We used a Split Decoder with 3 layers of hidden dimension 64 and Fourier features with a scale parameter of 5. The EpiNet architecture we used consisted of a trainable MLP with two hidden layers of dimension 32, and for the prior component an ensemble of 16 MLPs with 2 hidden layers of width 5 each and a scale parameter of 1. We trained this network for 4,000 steps using full batch and the Adam\cite{kingma2017adam} optimizer and exponential learning rate decay.

\begin{figure}[ht]
    \centering
    \includegraphics[width=0.49\textwidth]{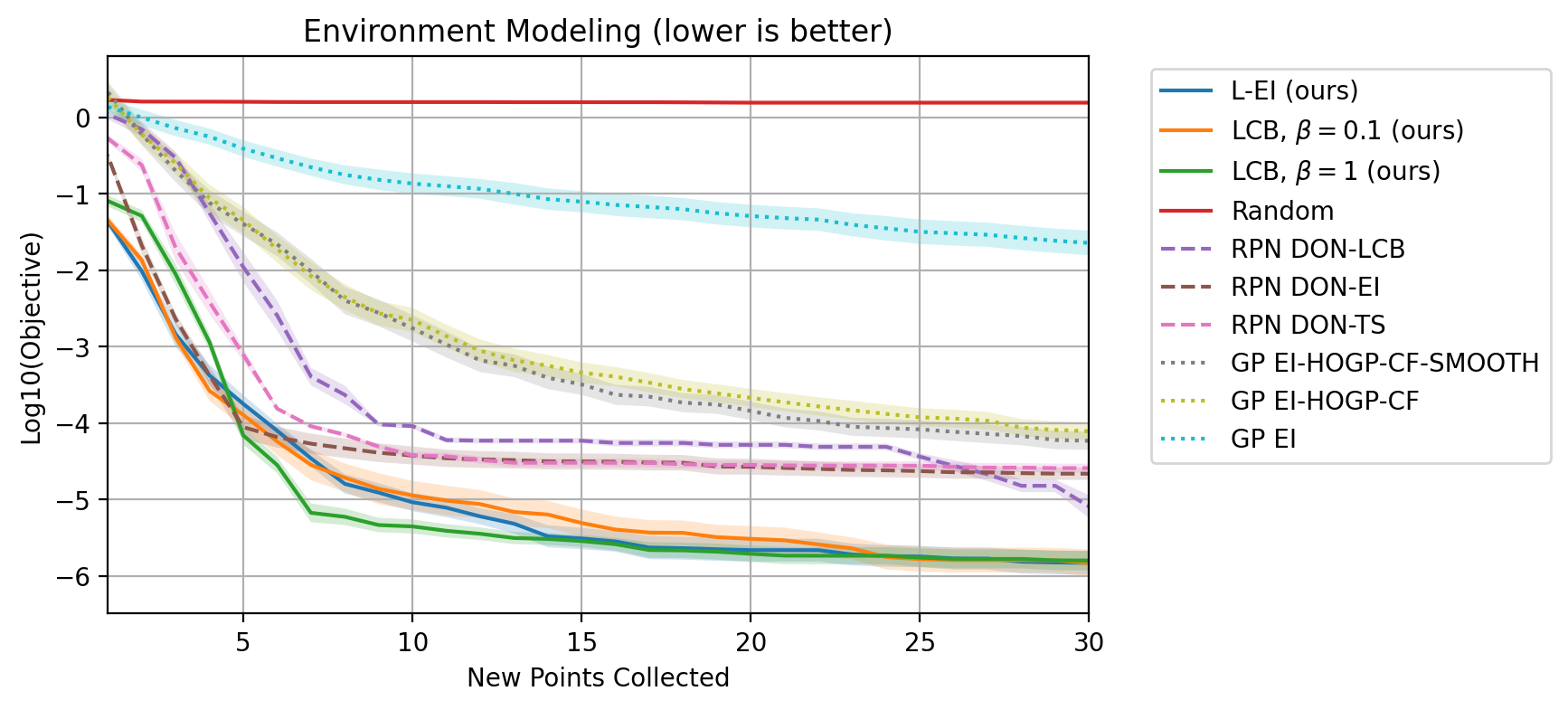}
    \includegraphics[width=0.49\textwidth]{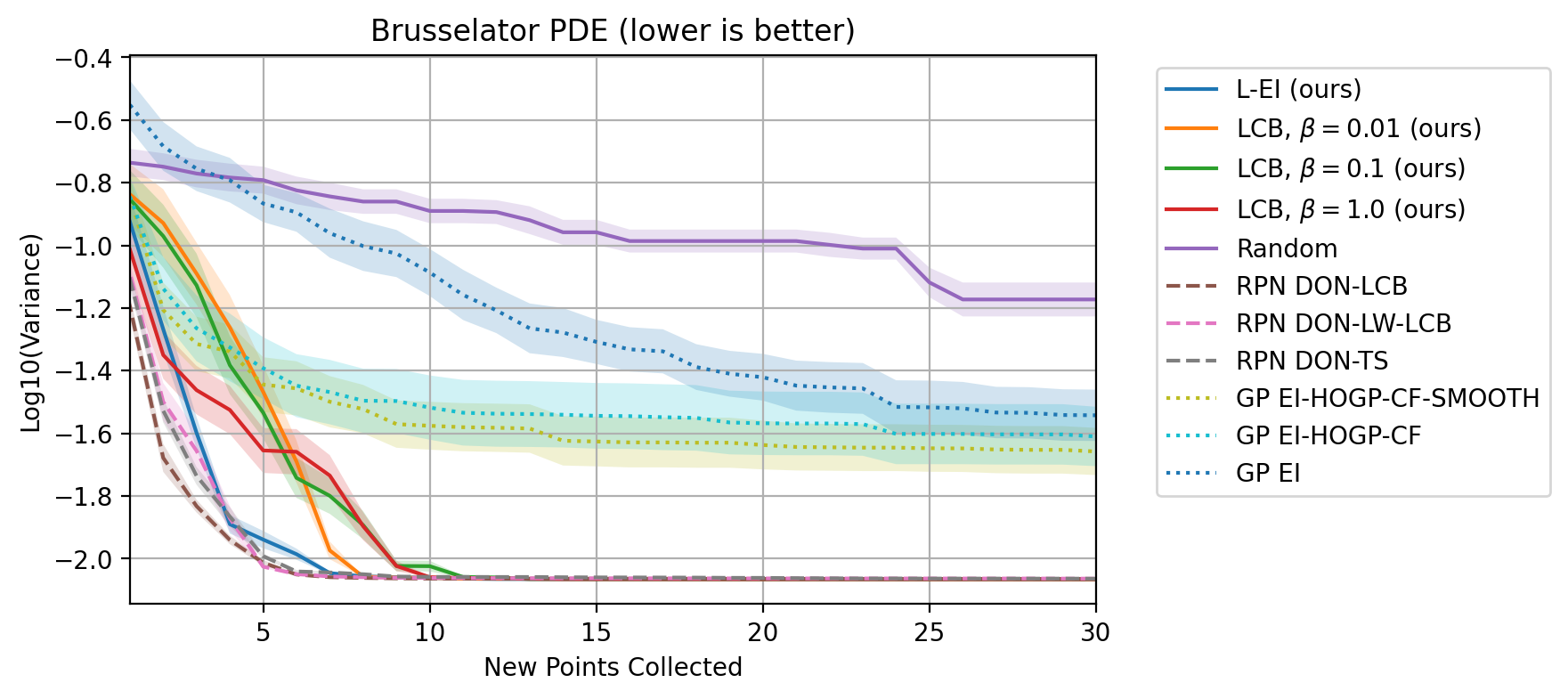}
    \caption{Full experimental results for the Environmental Modeling (left) and Brusselator PDE (right) problems. The dashed lines represent the RPN (Randomized Prior Network) ensemble reported in \cite{BHOURI2023116428} and dotted lines represent the Gaussian Process reported in \cite{maddox2021bayesian}. Following \cite{BHOURI2023116428}, the uncertainty bands indicate 20\% of the standard deviation band.}
    \label{fig:full_results_pollutants_brusselator}
\end{figure}

\subsectionPlus{Optical Interferometer}

This problem consists of optimizing the positional parameters of two mirrors in order to increase visibility. The input $u\in X=[-1,1]^4$ is used to compute the interference pattern, which is a set of 16 images on $[0,1]^2$, using a 64x64 grid. Thus, $h:X\to C([0,1]^2, \R^{16})$ is the unknown ground truth map, determined in our experiments using the InterferBot package\cite{sorokin2021interferobot}.

The objective function we aim to maximize is the visibility of $h(u)(x,y)=(I_1(x,y),\dots, I_{16}(x,y))$, defined by the map
\begin{align*}
    g(h(u)) = \frac{I_{\max}  - I_{\min}}{I_{\max}  + I_{\min}},
\end{align*}
where $I_{\max} = \text{LogSumExp}(\text{Intensity}_t)$ and $I_{\min} = -\text{LogSumExp}(-\text{Intensity}_t)$ and 
\begin{align*}
    \text{Intensity}_t = \int_0^1\int_0^1 \exp{\left(-(x-0.5)^2-(y-0.5)^2\right)} I_t(x,y) dxdy.
\end{align*}

\begin{figure}[h!t]
    \centering
    \includegraphics[width=0.85\textwidth]{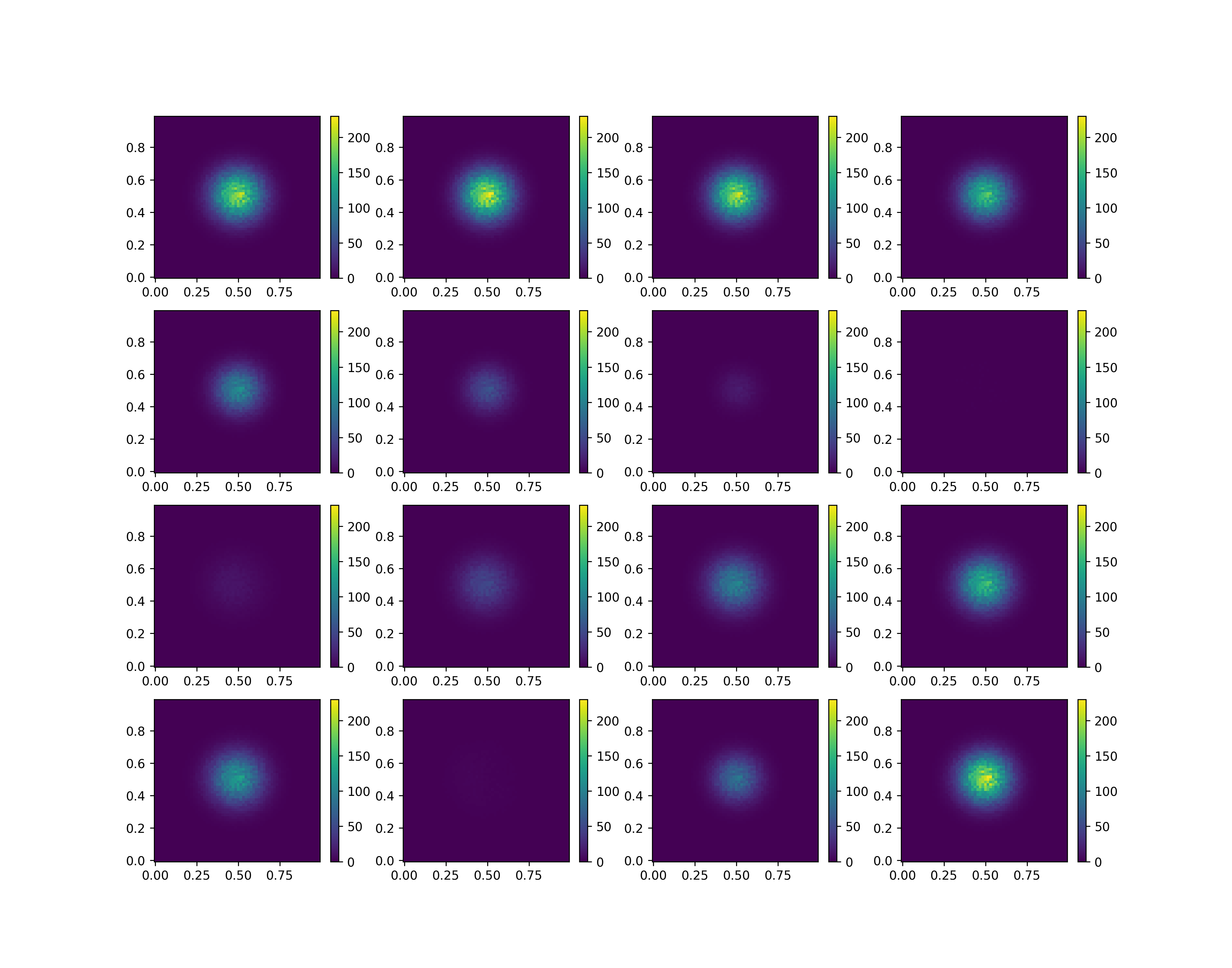}
    \caption{Best result obtained by NEON for the Optical Interferometer problem. Here we plot the 16 components of $h(u_\text{best})\in C([0,1]^2,\R^{16})$ where best input was $u_{\text{best}}=(-0.061, -0.0168,  0.121, -0.027)$, which yields an objective value of $f(u_{\text{best}})=g(h(u_{\text{best}}))=0.999$, very close to the upper possible limit of $1$.}
    \label{fig:best_interferometer}
\end{figure}

The NEON architecture for this experiment used an MLP encoder with 4 hidden layers with hidden dimension 64 and $d_\beta=96$. We used a Split Decoder with 6 layers of hidden dimension 128 and Fourier features with a scale parameter of 10. The EpiNet architecture we used consisted of a trainable MLP with 3 hidden layers of dimension 64, and for the prior component an ensemble of 16 MLPs with 2 hidden layers of width 8 each and a scale parameter of 1. We trained this network for 15,000 steps using a batch size of 16,384 and the Adam\cite{kingma2017adam} optimizer and linear warm up of the learning rate, followed by cosine decay.

\subsectionPlus{Cell Towers}

Given inputs $u\in X = (0,10)^{15}\times (30,50)^{15}\subset \R^{30}$ encoding antenna down-tilt angles and transmission strength, $h:X\to C([0,50]^2, \R^2)$ computes the signal strength and interference, respectively, of cellular service in the area. The objective $Obj$ we wish to optimize is then defined as the continuous version of the objective in \cite{maddox2021bayesian} as:
\begin{align*}
    Cov_{f,\text{strong}} &= \int\int \text{sigmoid}(T_w - R(x,y)) dxdy,\\
    Cov_{g,\text{weak, area}}(x,y) &= \text{sigmoid}\left(R(x,y)- T_w)\text{sigmoid}(I(x,y) + T_s - R(x,y)\right),\\
    Cov_{g,\text{weak}} &= \int\int \text{sigmoid}\left(I(x,y)Cov_{g,\text{weak, area}}(x,y) + T_w - R(x,y)Cov_{g,\text{weak, area}}(x,y)\right) dxdy,\\
    Obj &= 0.25Cov_{f,\text{strong}} + (1-0.25)Cov_{g,\text{weak}},
\end{align*}
where $h(u)(x,y) = (R(x,y), I(x,y))$ is evaluated on a 50x50 grid and $T_w=-80$ and $T_s = 6$ are the weak and strong coverage thresholds, respectively.

\begin{figure}[ht]
    \centering
    \includegraphics[width=0.7\textwidth]{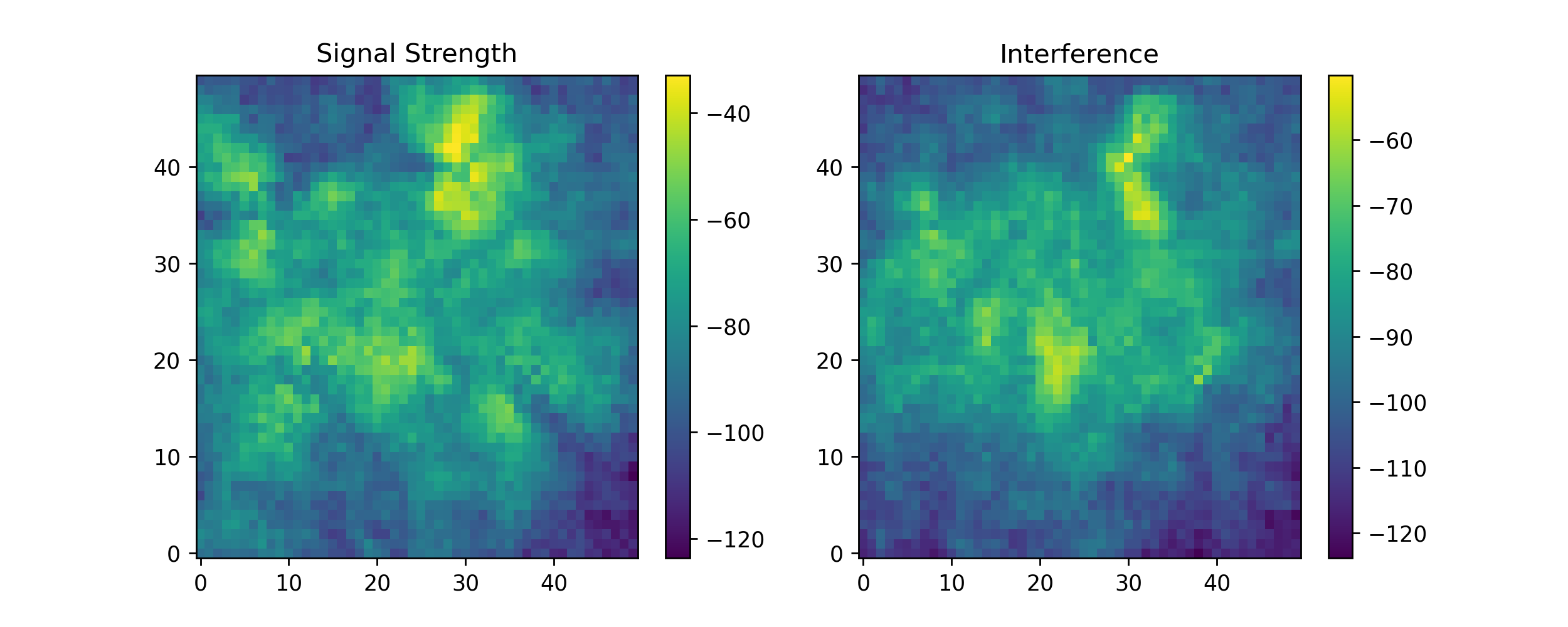}
    \caption{Best result obtained by NEON for the Cell Towers problem. Here we plot the signal strength and interference patterns $h(u_\text{best})\in C([0,1]^2,\R^{2})$, which yields an objective value of $f(u_{\text{best}})=g(h(u_{\text{best}}))=2171$.}
    \label{fig:best_cell_towers}
\end{figure}

The authors would like to point out the discrepancy between the objective function values we've obtained for this problem and those reported on \cite{maddox2021bayesian}. We have reached out to the authors of \cite{maddox2021bayesian} and they have indicated this arises from different versions of the simulation for the ground-truth function $h$, although the objective function $g$ used was identical. Since their version of $h$ is not publicly available, we used the simulator from \cite{dreifuerst2021optimizing} to compute the map $h$.

The NEON architecture for this experiment used an MLP encoder with 1 hidden layers with hidden dimension 64 and $d_\beta=192$. We used a Split Decoder with 6 layers of hidden dimension 64 and Fourier features with a scale parameter of 15. The EpiNet architecture we used consisted of a trainable MLP with 3 hidden layers of dimension 64, and for the prior component an ensemble of 16 MLPs with 2 hidden layers of width 5 each and a scale parameter of 0.5. We trained this network for 12,000 steps using a batch size of 2,500 and the Adam\cite{kingma2017adam} optimizer and linear warm up of the learning rate, followed by cosine decay.

\begin{figure}[ht]
    \centering
    \includegraphics[width=0.57\textwidth]{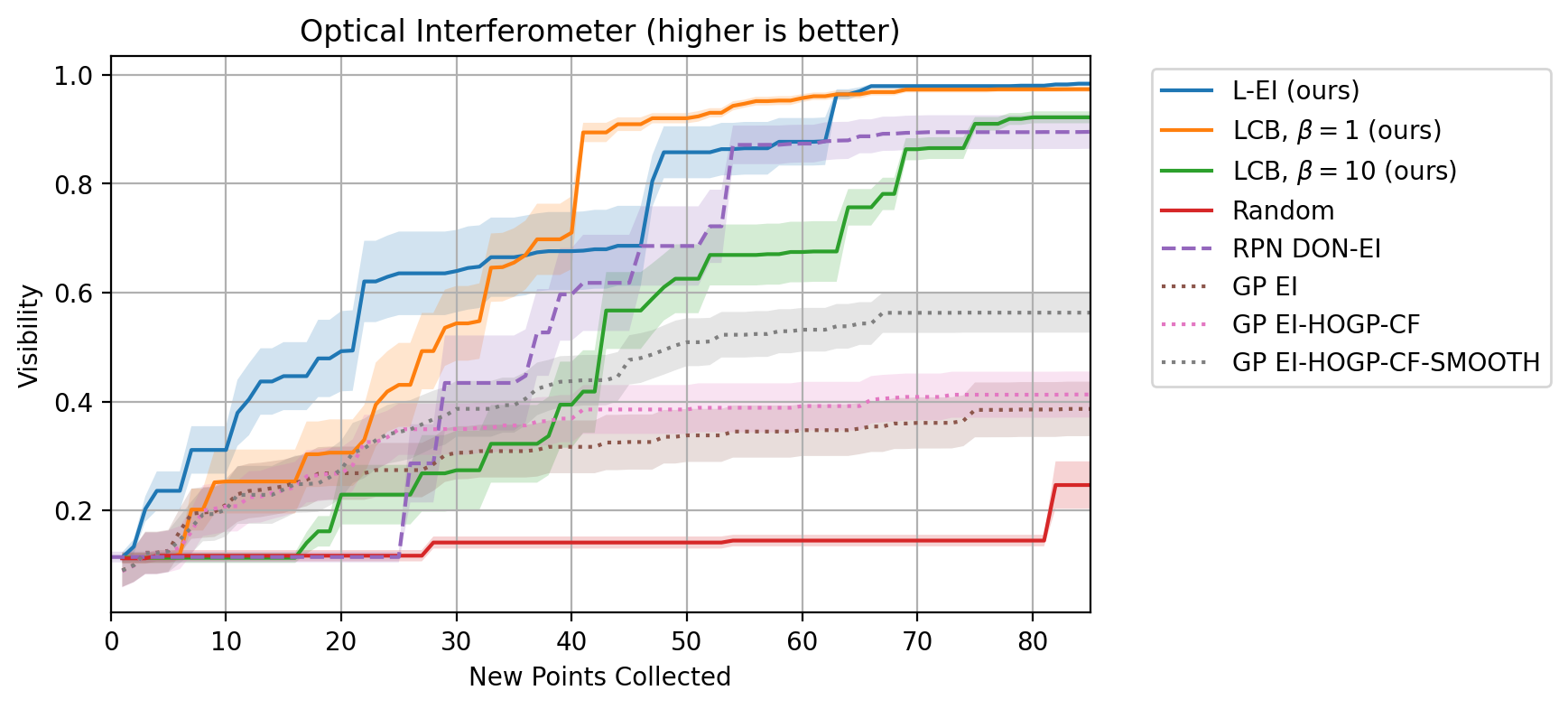}
    \includegraphics[width=0.42\textwidth]{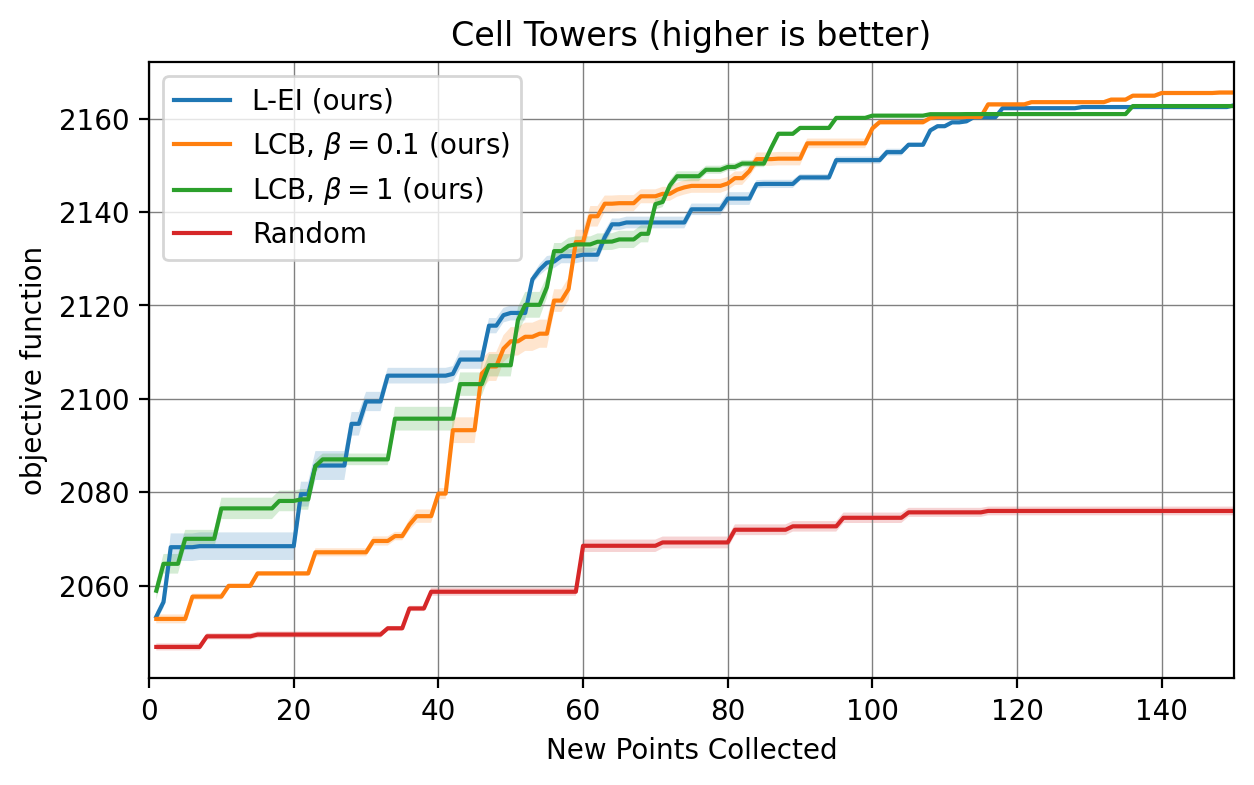}
    \caption{Full experimental results for the Optical Interferometer (left) and Cell Towers (right) problems. The dashed lines represent the RPN (Randomized Prior Network) ensemble reported in \cite{BHOURI2023116428} and dotted lines represent the Gaussian Process reported in \cite{maddox2021bayesian}. Following \cite{BHOURI2023116428}, the uncertainty bands indicate 20\% of the standard deviation band.}
    \label{fig:full_results_interferometer_cell}
\end{figure}

\sectionPlus{Parallel Acquisition}

In BO, it is possible to consider the problem of acquiring several new points at each iteration. This is done by selecting the points $x_1,\dots,x_q = \argmax \alpha_q(x_1,\dots, x_q)$ where $\alpha_q : X^q\to\R$ is a multi-point (or parallel) acquisition function. This may be desirable in situations where collecting a new data point is equally costly as acquiring $q>1$. For example, if an experiment requires harvesting cells from live animals, sacrificing a mouse may yield enough cells to perform several experiments simultaneously. In this case, the expensive component of evaluating the map $h$ is sacrificing the animal, but after doing so, performing a single experiment is approximately equal to performing $q>1$ ones. In addition to cases such as this, parallel acquisition also requires training surrogate models less times, as we only need to train one network in order to collect $q$ points at each iteration. Many single point acquisition functions have been extended to this parallel setting\cite{qEI,wilson2018maximizing}, and this framework is also compatible with NEON.

It has been shown that having quality measures of joint predictions is critical for decision making processes and that EpiNets excel at this task\cite{epistemicNNs}. This makes EpiNet based architectures such as NEON good candidates for parallel acquisition problems. We validate this claim by examining the performance of NEON in the Optical Interferometer problem using $q$-LEI, a generalization of the L-EI for $q\geq 1$ parallel acquisition analogous to $q$-EI\cite{qEI}. As such, $\alpha_{q\text{L-EI}}(x_1,\dots, x_q) = \mathbb{E}_z[\alpha'_{q\text{L-EI}}(x_1,\dots, x_q; z)]$ where
\begin{align*}
    \alpha'_{q\text{L-EI}}(x_1,\dots, x_q; z) =  \sum_{i=1}^q w_i(z)(G_\theta(x_i,z)-y_*),
\end{align*}
with
\begin{align*}
    w_i(z) = \begin{cases}
        1, & \text{if }G_\theta(q_i,z)=\max_{j=1,\dots,q}G_\theta(q_j,z)\text{ and }G_\theta(q_i,z)-y_*>0,\\
        0.01, & \text{else}.
    \end{cases}
\end{align*}
As was done in L-EI, the value $0.01$ can be substituted by any other $\delta>0$ if desired. This leads to an analogous statement to Theorem 1 for the parallel acquisition setting, with similar proof.

\begin{figure}[ht]
    \centering
    \includegraphics[width=0.4\textwidth]{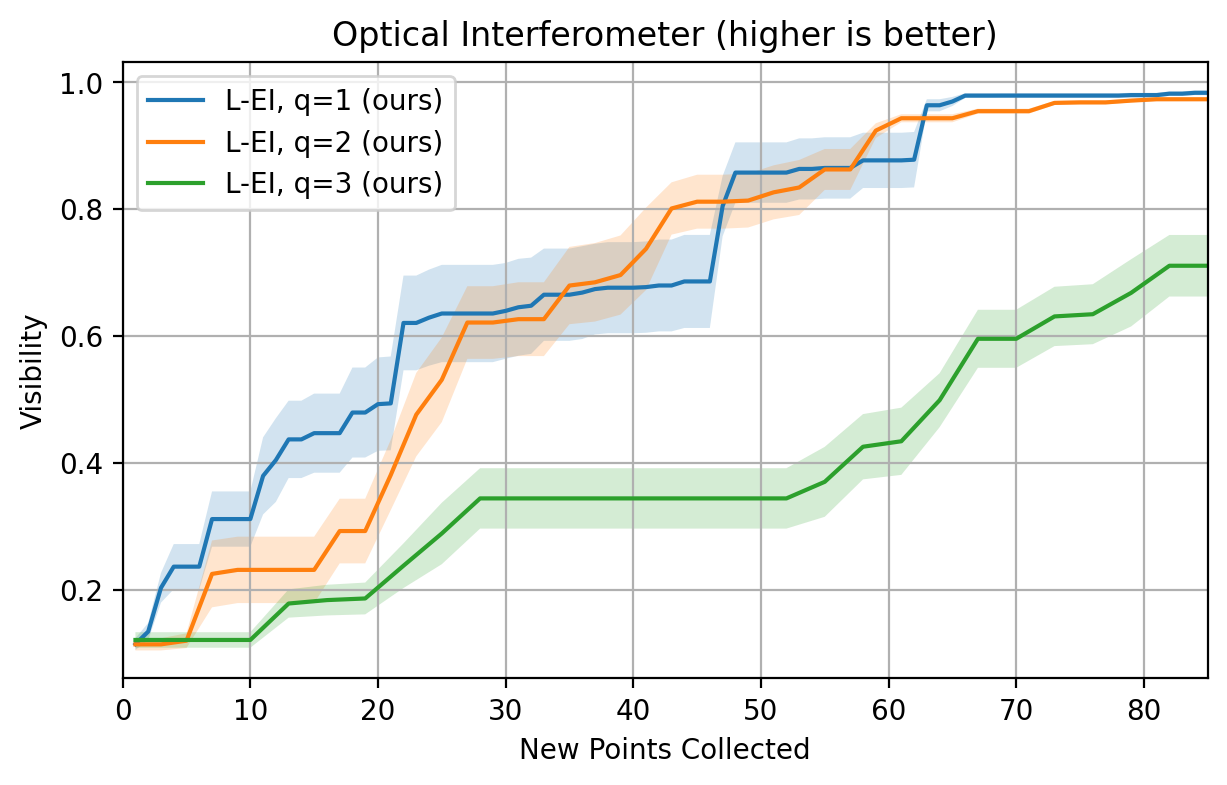}
    \includegraphics[width=0.4\textwidth]{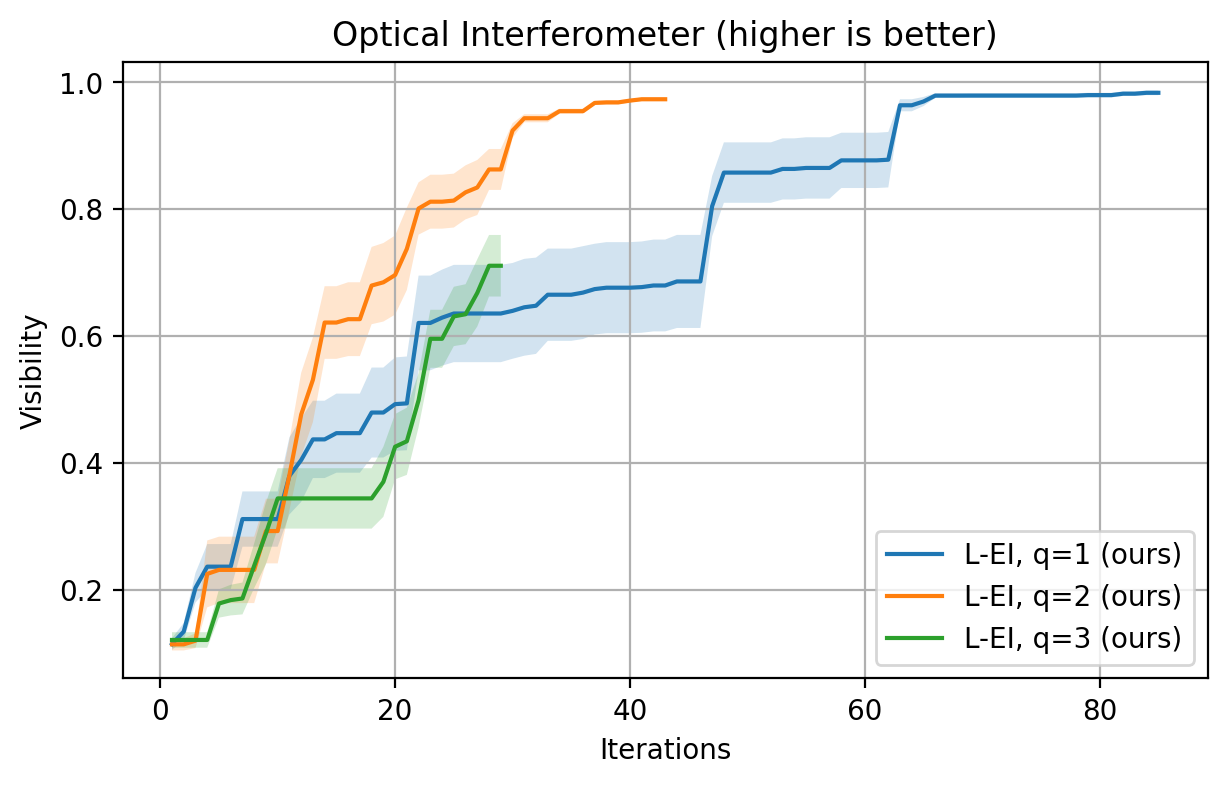}
     \caption{Average results among 5 trials comparing parallel acquisition functions using $q=2,3$ compared to the single-point version $q=1$. On the left, we compare the methods in terms of new points collected, while on the right we compare methods in terms of BO iterations. Following \cite{BHOURI2023116428}, the uncertainty bands indicate 20\% of the standard deviation band.}
    \label{fig:parallel_acquisition}
\end{figure}

The experimental results over 5 independent trials using different seeds are presented in Figure \ref{fig:parallel_acquisition}. As can be seen, using $q=2$ with the $q$-LEI acquisition yields similar results to the single-point acquisition function in terms of new points collected, and far superior in terms of BO iterations. Using $q=3$ yields slightly worse performance, but it is still able to optimize the objective function.

\end{document}